\documentclass[12pt]{article}

\usepackage{cite}
\usepackage{algorithm}
\usepackage{algorithmic}
\usepackage{latexsym,amsmath,amssymb,amsthm,amsfonts}
\usepackage[numbers,sort&compress]{natbib}
\usepackage{multirow}
\usepackage{array}
\usepackage{enumerate}
\usepackage{mathrsfs}
\usepackage{compactbib}
\theoremstyle{plain}
\usepackage{setspace}

\usepackage{graphicx,picture}
\usepackage{subfigure}
\usepackage{geometry} 
\usepackage{txfonts}

\newcommand{\tabincell}[2]{\begin{tabular}{@{}#1@{}}#2\end{tabular}}
\pdfoutput=1

\begin{document}

\title{Multi-view metric learning for multi-instance \\image classification}
\date{}
\author{Dewei~Li\thanks{Dewei Li is with the School of Mathematical Sciences, University
		of Chinese Academy of Sciences, Beijing 100049, China(E-mail: lidewei15@mails.ucas.ac.cn).} 
~and~Yingjie~Tian\thanks{Yingjie Tian is with the Research Center on Fictitious Economy and Data
		Science, Chinese Academy of Sciences; the Key Laboratory of Big Data
		Mining and Knowledge Management, Chinese Academy of Sciences, Beijing
		100190, China (Corresponding author. E-mail: tyj@ucas.ac.cn).}}

\maketitle

\begin{abstract}
It is critical and meaningful to make image classification since it can help human in image retrieval and recognition, object detection, etc. In this paper, three-sides efforts are made to accomplish the task. First, visual features with bag-of-words representation, not single vector, are extracted to characterize the image. To improve the performance, the idea of multi-view learning is implemented and three kinds of features are provided, each one corresponds to a single view. The information from three views is complementary to each other, which can be unified together. Then a new distance function is designed for bags by computing the weighted sum of the distances between instances. The technique of metric learning is explored to construct a data-dependent distance metric to measure the relationships between instances, meanwhile between bags and images, more accurately. Last, a novel approach, called MVML, is proposed, which optimizes the joint probability that every image is similar with its nearest image. MVML learns multiple distance metrics, each one models a single view, to unifies the information from multiple views. The method can be solved by alternate optimization iteratively. Gradient ascent and positive semi-definite projection are utilized in the iterations. Distance comparisons verified that the new bag distance function is prior to previous functions. In model evaluation, numerical experiments show that MVML with multiple views performs better than single view condition, which demonstrates that our model can assemble the complementary information efficiently and measure the distance between images more precisely. Experiments on influence of parameters and instance number validate the consistency of the method.
\end{abstract}


\section{Introduction}
Image classification and recognition has been an active research spot for many years with explosive growth of image data from daily life and internet. The far-reaching study can be employed in practical applications, including face recognition, species categorization, object detection, for a more intellective world. However, there exist threefold challenges in this research. The first is to extract numerical feature representation from images since original image cannot be exploited directly. Extensive studies have explored this area\cite{smeulders2000content}. Classical methods for feature extraction include Haar-like feature\cite{viola2001rapid,viola2004robust}, Scale-invariant feature transform(SIFT)\cite{lowe2004distinctive}, Histograms of Oriented Gradients(HOG)\cite{dalal2005histograms}, Local binary pattern(LBP)\cite{ahonen2004face}, Speeded up robust feature(SURF)\cite{bay2006surf}, etc. The above methods can be classified into two kinds: single feature vector representation(HAAR, HOG, LBP) and bag-of-words representation(SIFT, SURF, patches of LBP and HOG). The difference between the two kinds features is that a image is represented by a single vector or a bag of instances, leading to two areas in machine learning, standard classification and multi-instance learning. As that the contents in a image are not distributed uniformly or regularly,
bag-of-words representation has the advantage that every word expresses a image's key feature independently, without the negative impact of unfixed locations of these key features in a single vector. 
The second challenge is to integrate the information from multiple sources or multiple feature sets. In fact, the extracted features from different methods are complementary to each other and can be combined to improve the performance 
of image recognition, which is in the scope of multi-view learning. Multi-view learning has been developed from co-training, to multiple kernel learning and subspace learning\cite{xu2013survey}. Extensive experiments have verified that information form multiple views can boost the performance of methods in machine learning.
The third key challenge is to design an efficient data-dependent distance function to show the image relationships and distributions in feature space. The researchers in distance metric learning has proposed many algorithms to improve the performance of distance related methods based on the idea that a desired metric should shrink the distance between similar points and expand the distance between dissimilar points as much as possible\cite{kulis2012metric,moutafis2016overview,bellet2013survey}. 
The distance between images has been seldom studied when multi-instance features are extracted from images.
However, in practical, the features of image can be extracted from multiple views, each view consists of multi-instance features. It is worthy and important to unify the information from multiple views, investigating the relationship between different views and different bags in the same view.
It is much more interesting to explore multiple data-dependent metrics in multi-instance task with multiple views. 

In this paper, we propose a new approach to improve the performance of image classification, named MVML. For each image,  multi-instance features are extracted due to the merits of bag-of-words representation and different feature extraction methods are implemented to constructed multiple views. To combine the features from multi-view effectively, we first define a new distance function for bags and then seek for a instance-dependent metric by maximizing the average conditional probability that every image is similar with its nearest image. The distance between images is computed by the weighted sum of the distance between bags from each single view. So the metrics and weights are both needed to be optimized.
The tricks of gradient descent and alternate optimization are used to solve our approach. The efficiency of our novel method in making image classification, compared with single view multi-instance learning, has been demonstrated in the numerical experiments.
 
In summary, the main contributions of our work are as follows:
\begin{itemize}
	\item A new approach for multi-instance classification, based on multi-view learning with the technique of metric learning  is introduced. The combination of multi-instance, multi-view and metric learning has never been studied before. Multi-view learning have been explored to improve the performance of classification. 
	Metric learning can devise a view-dependent metric for every view to further boost the performance of multi-view learning.
	\item The proposed method is a conditional probability model, maximizing the average probability that every image is similar with its nearest image. In multi-view condition, the distance between images is the weighted sum of the distance between bags from each single view. Our method can be solved by gradient descent and alternate optimization iteratively.
	\item We have designed a new distance function between bags, integrating the distances between instances of bags skillfully. Compared with the previous two distances, experiments on $k$NN classification have validated the efficiency and advantage of our new distance function.
\end{itemize}

The followings of the paper are organized in this way. In Section \ref{sec:related}, we will introduce previous related works about metric learning, multi-view learning and multi-instance classification. Our model, including multi-instance problem with multi-view, distance function for bags and the probability framework will be provided in Section \ref{sec:model}. Then we will optimize our approach in this Section. In Section \ref{sec:exp}, numerical experiments will be made to demonstrate that our model can deal with multi-instance classification effectively and efficiently. The conclusions will be summarized in Section \ref{sec:conclu}.

\section{Related Works}
\label{sec:related}

\subsection{Metric learning}
Metric learning aims to learn a distance function to improve the performance of distance-related methods, including $k$-nearest neighbors($k$NN), $k$-means, which are very classical and important methods in classification, clustering, etc. For a dataset with $ c $ classes
\begin{equation}\label{tr01}
T = \{(x_1 ,y_1), \cdots ,(x_m ,y_m )\},
\end{equation}
where $(x_i,y_i) \in R^n \times \{ 1, 2, \cdots, c\}, i=1, \cdots, m$ and $ m $ is the total number of samples, $ n $ is the number of features. Two sets are defined as following
\begin{eqnarray}
S &=& \{(x_i,x_j) | y_i=y_j  \} \label{simpair} \\
D &=& \{(x_i,x_l) | y_i \neq y_l  \} \label{difpair}
\end{eqnarray}
The points in each pair of $ S $ are from the same class and $ D $ contains pairs of dissimilar points. Metric learning seeks for a metric to recompute the distance between two different points as
\begin{equation}\label{dist}
d_M(x_i,x_j)= (x_i -x_j)^\top M (x_i-x_j)
\end{equation}
to make similar points closer and dissimilar points farther. In the equation (\ref{dist}), an effective metric $ M $ should satisfies the conditions\cite{royden1988real,wang2014survey}:
\begin{enumerate}[(1)]
	\item distinguishability: $ d_M(x_i,x_i)=0 $;
	\item non-negativity: $ d_M(x_i,x_j) \ge 0 $;
	\item symmetry:  $ d_M(x_i,x_j)=d_M(x_j,x_i) $;
	\item triangular inequality: $ d_M(x_i,x_j)+d_M(x_i,x_k) \ge d_M(x_j,x_k)$;
\end{enumerate}

A great many methods of metric learning have been proposed to show the strong ability of data-dependent distance in adjusting the original structure of feature space, resulting in the formations of more advantageous neighborhoods. The label information of pairwise relationship in (\ref{simpair})-(\ref{difpair}) has been exploited in most of these previous works. One of the earliest efforts in pursuing ideal metric is the method MLSI(Metric Learning with Side Information), presented by Xing et.al\cite{xing2002distance}. It uses similarity side-information to improve $k$NN performance based on the idea that similar points should be as near as possible and the distance between dissimilar points should be larger than a threshold. Goldberger et.al proposes NCA(neighborhood component analysis)\cite{goldberger2004neighbourhood} which directly maximizes leave-one-out accuracy by learning a low-rank quadratic metric. Then LMNN(large margin nearest neighbor)\cite{weinberger2009distance} is presented on the basis of NCA to minimize the distance between any two similar and close points, subject to the constraints that any point should be pushed away from the neighborhood of its different labeled points by a large margin.
Due to the limitations of LMNN, several extensions are introduced to improve LMNN, including solving LMNN more efficiently\cite{park2011efficiently,weinberger2008fast}, introducing kernel into LMNN\cite{torresani2006large}, multi-task version of LMNN\cite{parameswaran2010large}, etc. Golberson has constructed a convex optimization problem\cite{globerson2005metric} to learn a quadratic Gaussian metric, trying to, though impractical, collapse all the similar inputs to a single point. From the view of information theory, ITML(information-theoretic metric learning)\cite{davis2007information} minimizes the relative entropy between two multivariate Gaussian distribution, leading to a Bregman optimization problem.

Metric learning for special task has also been studied extensively. SSM-DML\cite{yu2012semisupervised} pays attention to multi-view metric learning to improve the performance of cartoon synthesis. Jin et.al\cite{jin2009learning} propose an iterative metric learning algorithm for multi-instance and multi-label problem to improve the quality of associations between instances and class labels. MildML\cite{guillaumin2010multiple} learns a data-dependent metric of bags from multi-instance task based on the information that two bags in positive pair share at least one label, negative otherwise. MIMEL\cite{xu2011multi} aims to maximize inter-class bag distance and minimize intra-class bag distance, constructing a minimization problem of KL divergence between two multivariate Gaussians. 

\subsection{Multi-view learning}
To identify a person, the information of face, fingerprint or DNA can take effect independently. These three kinds of information are from different sources. The features of color, shape or texture can be used to classify pictures individually. They can be seen as different feature subsets of the images. What's more, we can also combine features from different sources or feature subsets to identify objects with higher accuracy since diverse characters are synthesized. It is the basic idea of multi-view learning. Each source or subset is called a view. 
In contrast to single view learning, multi-view learning can improve the learning performance by optimizing multiple functions, each one is used to model a single view, to extract the information from different views of the same data inputs. Multi-view learning has been a hot research spot recently and the related studies can be divided into three directions: co-training, multiple-kernel learning and subspace learning\cite{xu2013survey,sun2013survey}. The first co-training algorithm was proposed for semi-supervised learning\cite{blum1998combining}, which combines the information from two views, labeled and unlabeled, to boost the performance of the algorithms with only few labeled data points. It had been analyzed and extended further\cite{nigam2000analyzing,muslea2002active+,yu2011bayesian}. Kernel function is an important and useful tool in machine learning, which maps original example into higher dimensional Hilbert space for easier learning with better performance. Multiple kernel learning(MKL) can be applied into multi-view learning task, each kernel corresponds to a particular view. Multiple kernels will be combined linearly or non-linearly to model multi-view data with more proper way. MKL had been studied in diverse formations, including semi-definite program\cite{lanckriet2004learning,sonnenburg2005general}, second order cone program\cite{bach2004multiple}, 2-norm regularization program\cite{rakotomamonjy2007more}, etc and explored widely\cite{ying2009generalization,kloft2011local}. 
Canonical correlation analysis(CCA)\cite{hotelling1936relations} and its kernel version\cite{akaho2006kernel} are two early works in subspace learning. They seek basis vectors for two sets of variables, each set can be seemed as the data corresponds to a single view. They have been extended to clustering\cite{chaudhuri2009multi} and regression\cite{diethe2008multiview} and analyzed for consistency\cite{cai2011convergence}.

\subsection{Multi-instance learning}
Different from standard supervised learning, in which the input is often described by a single feature vector, 
every input in multi-instance learning(MIL) is a set of labeled instances, called bag. The label of a bag is determined by one or several instances. The multi-instance problem was first introduced by Dietterich\cite{dietterich1997solving} in the study of drug activity prediction. Drug contains many molecules, valid and invalid. If the molecules in a drag are all invalid, the drug is labeled as negative, other wise positive. Classification or pattern recognition of multi-instance task can be considered on bag level and/or instance level. APR(Axis-Parallel Rectangle)\cite{dietterich1997solving} is an early work in dealing with MIL, starting from a initial point and expanding a rectangle to find the smallest rectangle that covers at least one instance of each positive bag and zero instance of any negative bag. A probabilistic framework called Diverse Density(DD)\cite{maron1998framework} was proposed to learn a concept by maximizing a defined likelihood function. Zhang et.al\cite{zhang2001dd} proposes an improved version of DD which combines Expectation-Maximization and diverse density. Citation-$k$NN\cite{wang2000solving} is introduced to decide the label of a bag not only by its neighbors but also citers. Support vector machine had been adopted in MIL with modified constraints that some instances are unlabeled but should be constrained\cite{andrews2002support,tao2004svm}.
 
However, there are few research on multi-instance problem with multiple views, let alone the technique of metric learning embedded. MVML is a probability framework, defining a distance function for multi-instance problem and integrating the information from multiple views by learning multiple metrics. The experiments verifies that our approach is effective in dealing with multi-instance task.

\section{Model and optimization}
\label{sec:model}
\subsection{Multi-instance classification task with multi-view }
In multi-instance task, the features of every image are extracted in the form of bags, each of which contains multiple instances. Under the condition of multi-view, every image can be described by $ v $ different but complementary views. We can classify these images based on a single view  or multiple views. Generally, multi-view data can provide more information.
To make classification on the batch of images, the training set with $ v $ views is constructed as following:
The $k$-th($ k=1, 2, \cdots, v$) view of the training set is
\begin{equation}\label{eq:prbk}
T^k=\{(X_1^k,y_1), (X_2^k,y_2), \cdots, (X_{m_k}^k,y_{m_k})\}  
\end{equation}
where $ X_i^k=\{x_{i1}^k, x_{i2}^k, \cdots, x_{i m_i^k}^k \}, y_{j_k} \in \{1,2, \cdots, c\} $. $ T^k $ is composed of $ m_k $ bags with their corresponding labels. The bag $ X_i^k $ contains $ m_i^k $ instances. 
It should be noted that the above mentioned problem is not the same as the traditional multi-instance task, in which the label of a bag is determined by whether the bag contains positive instance. 
In our problem, the classes $ c $ is equal with or larger than 2 and all the instances in a bag may be helpful in deciding the label.

The task of the paper is to find a prediction function $ f $, with respect to multiple distance metrics $ M_1, \cdots, M_v $, each corresponds to an unique view: given an image with the information from $ v $ views, $ X^1, X^2, \cdots, X^v $, the label of the image can be predicted by $ y=f(X^1, X^2, \cdots, X^v; M_1, M_2, \cdots, M_k) $.

\subsection{Distance between Bags}
\label{sec:dist}
In the traditional distance metric learning, the distance is measured between two feature vectors $ x_i. x_j $, that is,
\begin{equation}
d_M(x_i, x_j)=(x_i-x_j)^\top M(x_i-x_j)
\end{equation}
where the metric $ M $ should satisfies the property of distinguishability, non-negativity, symmetry, triangular inequality. However, in the feature extraction of image, the feature is often in the form of bag, which contains multiple vectors. It is not suitable to concatenate these vectors into a longer vector. So it is hard but very important to measure the distance of different images, each of which is represented by a bag of features. 

In metric learning, there are two distance functions for bags proposed before, both of which are designed to measure the distance between bags as accurately as possible. The first one is to measure the distance by the average distance of pairwise examples from different bags\cite{xu2011multi}.
The second one calculates the minimum distance of pairwise instances\cite{jin2009learning}, also called \textit{minimal Hausdorff distance}. For two bags $ X_i=\{x_{i1}, x_{i2}, \cdots, x_{im_{i}}\}, X_j=\{x_{j1}, x_{j2}, \cdots, x_{jm_{j}}\} $, the distances defined by the above two functions are 
\begin{equation}
D_{ave}(X_i,X_j;M)=\frac{1}{m_i m_j} \sum\limits_{k=1}^{m_i} \sum\limits_{l=1}^{m_j} d_M(x_{ik}, x_{jl})
\end{equation}
and
\begin{equation}
D_{min}(X_i,X_j;M)= \min\limits_{1 \le k \le m_i, 1 \le l \le m_j} d_M(x_{ik}, x_{jl})
\end{equation}
respectively.
However, the two functions can be improved in virtue of their own drawbacks. For the function $ D_{ave} $, calculating the distance of pairwise instances can bring the trouble of much redundant information. The distance between two same bags is not zero in $ D_{ave} $. For the function $ D_{min} $, it determines the distance of bags only by the minimum distance of pairwise instances, which ignores too much useful information. If two different bags contain similar instances, $ D_{min} $ will make improper judge. So we argue that both $ D_{ave}$ and $D_{min} $ can not measure the bag distance properly. 

In the paper, a new distance function is defined as follows:
for each instance $ x $ in each bag, the nearest instance in the other bag is found and the distance is recorded. The distance is the minimum for $ x $ in searching the other bag. The new defined distance function calculates the Average of these Minimums, named as $ D_{am} $.
We define the distance between two bags $ X_i, X_j $ as 
\begin{equation}
D_{am} (X_i, X_j; M) =\frac{1}{m_i} \sum\limits_{p=1}^{m_i} \min\limits_{x_{jl} \in X_j} d_M(x_{ip}, x_{jl}) + \frac{1}{m_j} \sum\limits_{q=1}^{m_j} \min\limits_{x_{ih} \in X_i} d_M(x_{jq}, x_{ih})
\end{equation}

\begin{figure}
	\centering
	\subfigure[]{\includegraphics[width=3.5in]{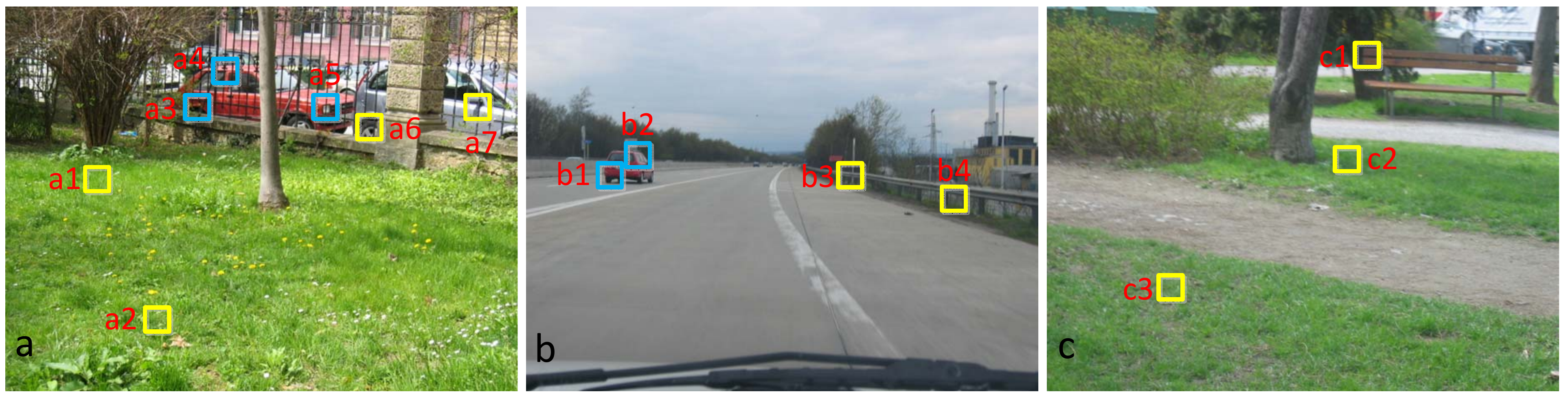}}	
	\subfigure[]{\includegraphics[width=2.5in]{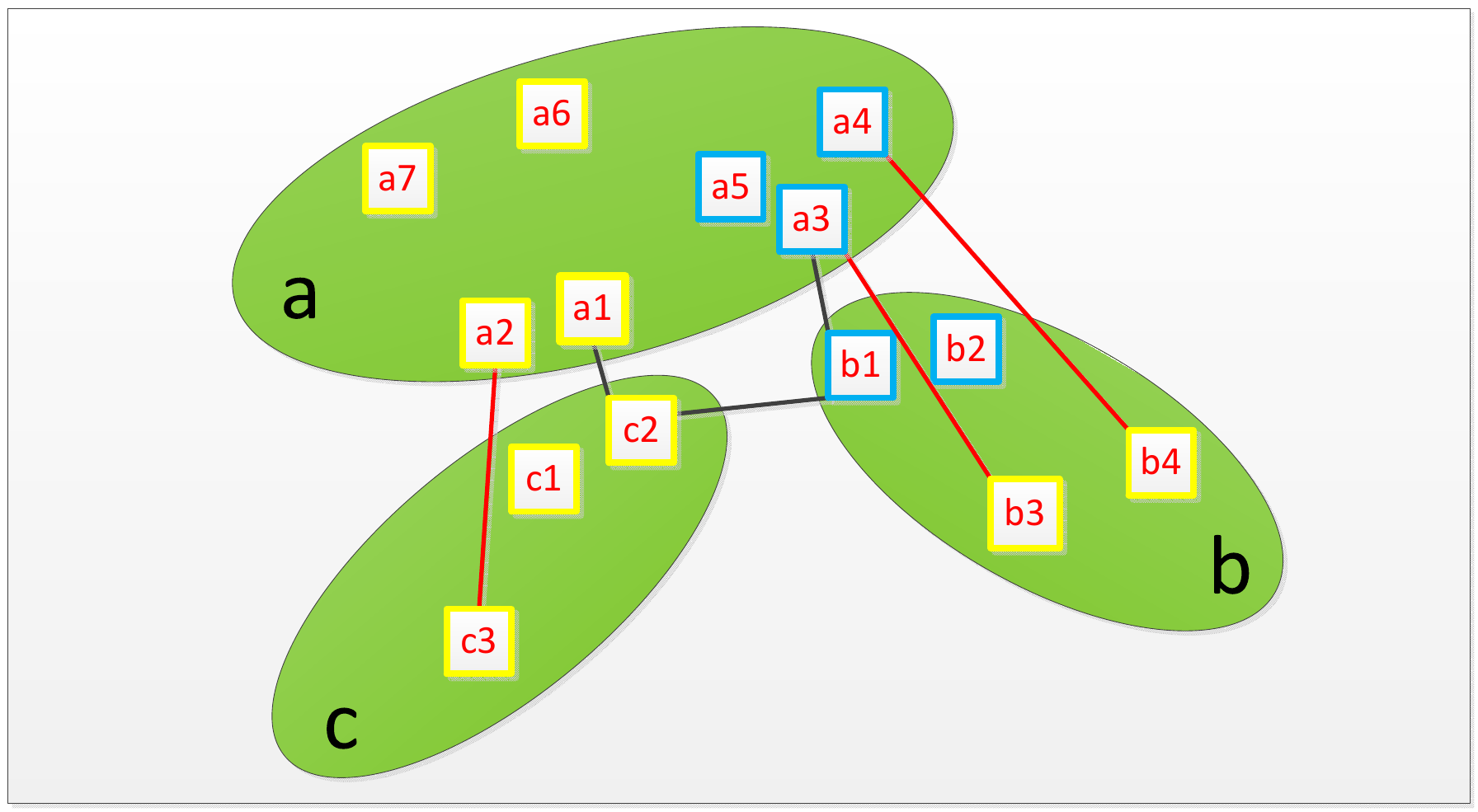}}
	\caption{A simple task}\label{fig:distconcept}
\end{figure}
Suppose that there is a simple task in Figure \ref{fig:distconcept}: Give a picture \textbf{a} belonged to the class of \textit{car}, find the picture with the same label from \textbf{b} and \textbf{c}. There are 7, 4, 3 key points in the picture \textbf{a, b, c} respectively. The blue squares are the key points indicating the label. On the instance level, the key points b1,b2 in b are similar with a3, a4 and c2, c3 are similar with a1, a2. In the function $ D_{min} $, the distances between pictures are $ D_{min}(a,b)=d(a3,b1) $ and $ D_{min}(a,c)=d(a1,c2) $(The black lines in Figure \ref{fig:distconcept}.(b) are pairwise distances calculated by $D_{min}$), both are relative small. We will make wrong judge if the latter is a slight smaller than the former one. So similar instances from different classes can weaken the quality of $ D_{min} $. In the function $ D_{ave} $, all the distances of pairwise instances will be computed and averaged, bringing negative information, such as $ d(a3,b3), d(a4,b4), d(a3,b4) $, etc, which will not be calculated in $ D_{am} $(In Figure \ref{fig:distconcept}.(b), the red lines are redundant distance information in computing bag distance). The experiments in the Section \ref{sec:exp} will verify the advantages of our proposed distance function.

In multi-view condition, every image $ I $ is extracted as $ v $ bags $ X_i^1, X_i^2, \cdots, X_i^v $. The distance between two images $ I, J $ is defined as
\begin{equation}
D_M(I, J)=\sum\limits_{k=1}^v \alpha_k D_{am}(X_i^k, X_j^k)
\end{equation}
where $ \alpha_i, i=1,\cdots, v $ are the weights to be learned.

\subsection{Metric learning in probability framework}
Although the distance between bags is designed, the relationship between instances is also very important since it affects bag distance significantly. Fortunately, metric learning can be applied to establish favorable relationships for instances form bag.
Inspired by the ideas of $k$NN and NCA, for each image, we just aim to maximize the probability of that its nearest image has the same label with it. 
In multi-view situation, we consider optimizing the joint conditional probability distribution of the image $ I $ with $ v $ views, but not simply maximizing the sum of the marginal probability distribution, 
\begin{equation}\label{mlp01}
p(y_i|X_i^1, X_i^2, \cdots, X_i^v;\mathscr{M}) =\frac{\exp(-f(X_i,y_i))}{ \sum\limits_{y=1}^c \exp(-f(X_i,y))}
\end{equation}
where 
\begin{equation}\label{mlp02}
f(X_i, y) =\min\limits_{y_j=y} D_M(I,J)= \min\limits_{y_j=y} \sum\limits_{k=1}^v \alpha_k D_{am} (X_i^k, X_j^k)
\end{equation}
and $ \mathscr{M}=\{M_1, \cdots, M_v\} $.

On the whole training set, the following regularized likelihood function is constructed to learn the desired distance metrics
\begin{eqnarray}
&& \max\limits_{\mathscr{M},\alpha}  E(\mathscr{M},\alpha) \nonumber \\
&=& \frac{1}{m} \sum\limits_{i=1}^m \ln p(y_i|X_i^1, X_i^2, \cdots, X_i^v) \nonumber - \frac{\lambda}{2} \sum\limits_{k=1}^v \|M_k\|_F^2 -\frac{\mu}{2} \|\alpha\|^2 \\
&=& -\frac{1}{m} \sum\limits_{i=1}^m f(X_i,y_i)- \frac{1}{m} \sum\limits_{i=1}^m \ln \sum\limits_{y=1}^c \exp(- f(X_i,y)) \nonumber \\
&& - \frac{\lambda}{2} \sum\limits_{k=1}^v \|M_k\|_F^2 -\frac{\mu}{2} \|\alpha\|^2  \label{model01}
\end{eqnarray}
To ensure the basic property of metric, $E(M,\alpha)$ should be subject to $M_k \succeq 0, k=1,\cdots,v$(positive semi-definite).

Let $ v=1 $, the primary model degenerates into a single view version and the objective function (\ref{model01}) becomes
\begin{equation}
 \max\limits_{M,\alpha}  E_s(M,\alpha) 
= \frac{1}{m} \sum\limits_{i=1}^m \ln p(y_i|X_i) \nonumber - \frac{\lambda}{2} \|M\|_F^2  \label{model02}
\end{equation}
which can be used to learn metric in multi-instance classification with single view.

\subsection{Optimization}
\label{sec:opt}
Although the model is constructed with the constraints that all the metrics should be positive semi-definite,
we can first optimize the model in an unconstrained condition and get $ M_1^*, M_2^*,\cdots, M_v^* $ and then project the metrics into positive semi-definite space. The procedure proceeds iteratively until convergence.
Our algorithm can be solved by gradient ascent and projection alternately. 

In single view version, only one metric need to be optimized. The model can be solved in the Algorithm \ref{alg:svmiml}. Given an unknown image \textit{L} with bag $ X_l $, its label can be decided by
\begin{equation}
y=\arg \max\limits_{y=1,\cdots,c} p(y_l|X_l; M^*)
\end{equation}
\begin{algorithm}
	\caption{Single view metric learning for multi-instance task (SVML)}\label{alg:svmiml}
	\textbf{Input:} The training set $ T^1$; The penalty parameters $ \lambda$, gradient step-size $ \eta $, maximum of iterations $ R $.\\
	\textbf{Output:} The target metrics $ M^*$;\\
	\textbf{Procedure:}\\
	1. Let $ r=1$ and initialize $ M$ as identity matrix;\\
	2. Update $ M $ alternately using gradient ascent method by
	\begin{equation}
	M^{\text{new}}=M^{\text{old}}+\eta \frac{\partial E_s}{\partial M}
	\end{equation}
	and make projections by
	\begin{equation}
	M^{\text{new}}=\text{PSD}(M^{\text{new}})
	\end{equation}
	where PSD denotes the projection operator of positive semi-definite space.\\
	3. Let $ r=r+1 $, if $ r > R $, stop iteration and obtain the output, otherwise go to step 2.\\
\end{algorithm}

First, fix $\alpha_k, k=1,\cdots,v$, the gradient of $ E(M,\alpha) $ with respect to $ M_k $ is
\begin{eqnarray}
\frac{\partial E}{\partial M_k} &=& -\frac{1}{m} \sum\limits_{i=1}^m \frac{\partial f(X_i,y_i)}{\partial M_k} - \lambda M_k \nonumber \\
&& + \frac{1}{m} \sum\limits_{i=1}^m \frac{\sum\limits_{y=1}^c \exp(-f(X_i,y)) \frac{\partial f(X_i,y)}{\partial M_k}}{\sum\limits_{y=1}^c \exp(-f(X_i,y))}
\end{eqnarray}
To obtain $\frac{\partial f(X_i,y)}{\partial M_k}$, we decompose it as
\begin{eqnarray}
\frac{\partial f(X_i,y)}{\partial M_k} &=& \frac{\partial }{\partial M_k} \alpha_k^* D_{am} (X_i^k, X_{j^*}^k) \\
&=& \frac{\alpha_k^*}{m_i^k} \sum\limits_{p=1}^{m_i^k} \frac{\partial }{\partial M_k} \min\limits_{x_{jl} \in X_j} d_M(x_{ip}, x_{jl}) \nonumber \\
&& + \frac{\alpha_k^*}{m_j^k} \sum\limits_{q=1}^{m_j^k} \frac{\partial }{\partial M_k} \min\limits_{x_{ih} \in X_i} d_M(x_{jq}, x_{ih})\\
&=& \frac{\alpha_k^*}{m_i^k} \sum\limits_{p=1}^{m_i^k} (x_{ip}-x_{jl^*})(x_{ip}-x_{jl^*})^\top \nonumber \\
&& +\frac{\alpha_k^*}{m_j^k} \sum\limits_{q=1}^{m_j^k} (x_{jq}-x_{ih^*})(x_{jq}-x_{ih^*})^\top \label{eq:fpar}
\end{eqnarray}
where
\begin{eqnarray}
(\alpha_k^*, j^*)=\arg\min\limits_{\alpha,j} \sum\limits_{k=1}^v \alpha_k D_{am} (X_i^k, X_j^k)
\end{eqnarray}
For each $ p=1,\cdots, m_i^k $, 
\begin{equation}
l^*=\arg \min\limits_{x_{jl} \in X_j} d_M(x_{ip}, x_{jl})
\end{equation}
and for each $ q=1,\cdots, m_j^k $,
\begin{equation}
h^*=\arg \min\limits_{x_{ih} \in X_i} d_M(x_{jq}, x_{ih})
\end{equation}

Second, fix $ M_k, k=1,\cdots,v $, update $ \alpha $ by 
$ \alpha= \alpha+ \eta_2 \frac{\partial E}{\partial \alpha} $, where
\begin{eqnarray}
\frac{\partial E}{\partial \alpha} &=& -\frac{1}{m} \sum\limits_{i=1}^m F(X_i,y_i) -\mu \alpha \nonumber \\
&& + \frac{1}{m} \sum\limits_{i=1}^m \frac{\sum\limits_{y=1}^c \exp(- f(X_i,y)) F(X_i,y)}{\sum\limits_{y=1}^c \exp(-f(X_i,y))}
\end{eqnarray}
where 
\begin{equation}
F(X_i,y)=(D(X_i^1,X_j^1), \cdots, D(X_i^v,X_j^v))^\top 
\end{equation}
and 
\begin{equation}
X_j=\arg \min\limits_{y_j=y} \sum\limits_{k=1}^v \alpha_k D_M (X_i^k, X_j^k)
\end{equation} .
It should be noted that the equation (\ref{eq:fpar}) is computed on the basis of the metrics and weights, which are unknown before the equation. So iterative optimization will be implemented in our method. The metrics and weights for different views will be first initialized and then $ M_k, \alpha_k, k=1,\cdots,v $ can be updated by gradient ascent and projection to the space of positive semi-definite iteratively until convergence.
The detailed procedure of our method is shown in the Algorithm \ref{alg:mvmiml}.
Given an unknown image \textit{L} with $ v $ bags $ X_l^1, \cdots, X_l^v $, its label can be decided by
\begin{equation}
y=\arg \max\limits_{y=1,\cdots,c} p(y_l|X_l^1, \cdots, X_l^v; \mathscr{M}^*)
\end{equation}
where $ \mathscr{M}^*=\{M_1^*, \cdots, M_v^*\} $.
\begin{algorithm}\label{algo:mvml}
\caption{Multi-view metric learning for multi-instance task (MVML)}\label{alg:mvmiml}
\textbf{Input:} The training set $ T^1, T^2, \cdots, T^v $; The penalty parameters $ \lambda, \mu $, gradient step-size $ \eta $, maximum of iterations $\tau, R $.\\
\textbf{Output:} The target metrics $ M_1^*, M_2^*, \cdots, M_v^* $;\\
\textbf{Procedure:}\\
1. Let $ t=1, r=1 $ and Initialize $ M_1, M_2, \cdots, M_v $ as identity matrix;\\
2. Fix $ \alpha_k, k=1, 2, \cdots, v $, update $ M_k, k=1,\cdots,v $ alternately using gradient ascent method by
\begin{equation}
M_k^{\text{new}}=M_k^{\text{old}}+\eta_1 \frac{\partial E}{\partial M_k}
\end{equation}
and make projections to positive semi-definite space
\begin{equation}
M_k^{\text{new}}=\text{PSD}(M_k^{\text{new}})
\end{equation}
3. Let $ t=t+1 $, if $ t > \tau $, stop iteration, otherwise go to step 2;\\
4. Update the weight $ \alpha $,
\begin{equation}
\alpha^{\text{new}}= \alpha^{\text{old}}+\eta_2 \frac{\partial E}{\partial \alpha}
\end{equation}
5. Let $t=1, r=r+1 $, if $ r > R $, stop iteration and obtain the output, otherwise go to step 2.\\
\end{algorithm}

\begin{figure*}
	\centering
	\includegraphics[width=7in]{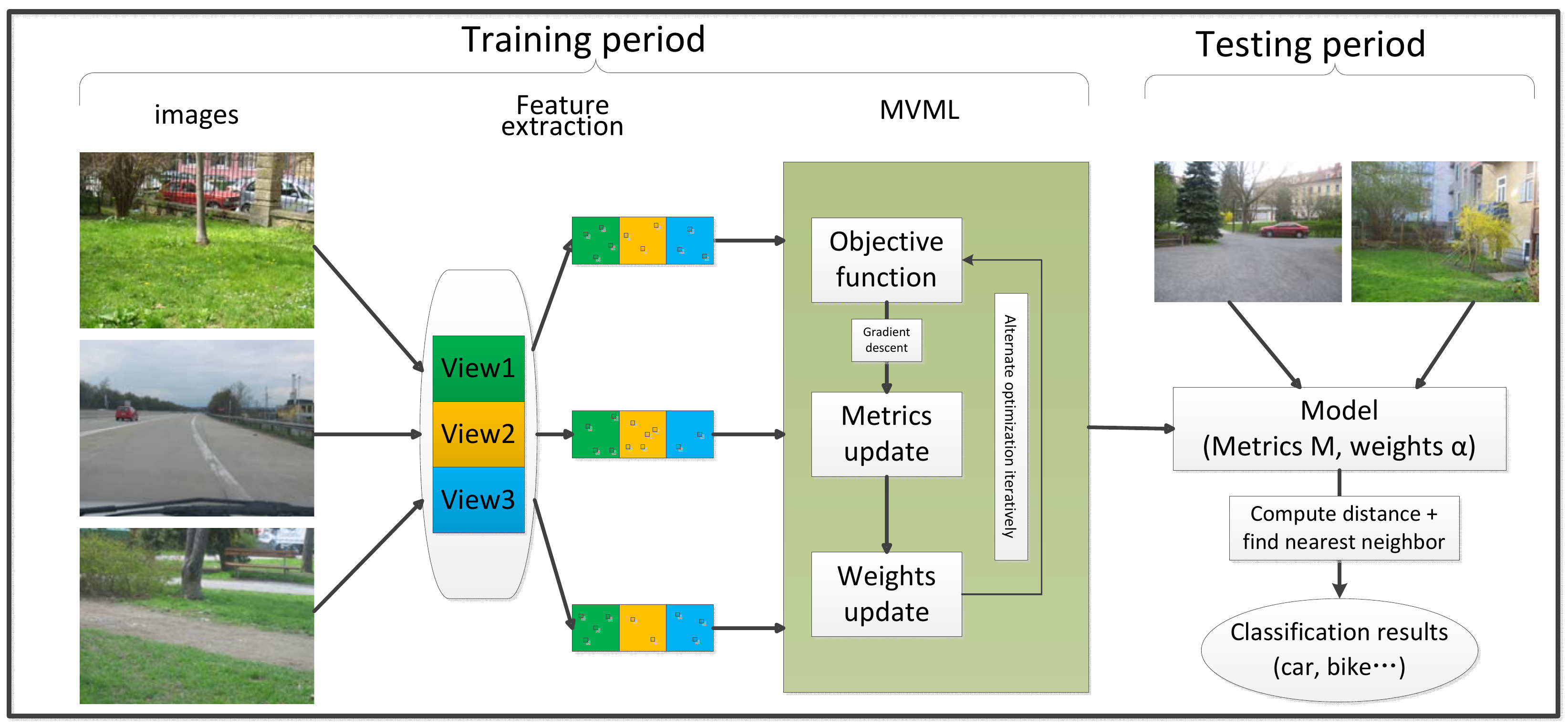}
	\caption{Workflow}\label{fig:work}
\end{figure*}

\subsection{Computational cost}
Our approach is solved iteratively with alternate optimization. The computation of gradient descent is the main part of the computational cost. It contains two parts: updating metrics and weights. In updating the metrics, the computational cost in every iteration is $O(V^2 K^2 mn(m+n)) $, where $ V $ is the number of views, $ K $ is the number of images, $ m $ is the average number of instances in each bag and $n$ is the average length of each instance. And the cost is $O(V K^2 mn(m+n)) $ in updating weights. In consideration of the iteration number $ R, \tau $. The total computational cost of our model is $ O(R\tau V^2 K^2 mn(m+n))$. The cost is relatively high, but tractable in experiments, which can be improved by parallel computing or GPU acceleration in the future.

\section{Experiments}
\label{sec:exp}
In this section, numerical experiments on image datasets are made to demonstrate the efficacy of our new method. Six datasets Corel, Caltech, GRAZ02(people, cars, bikes, background), Butterfly, Galaxy Zoo, FERET are selected and converted into standard format(Each dataset contains three parts: features, bag ids and labels). 
All the experiments are made on Matlab 2015a(PC, 8GB RAM).

\subsection{Datasets}
\begin{figure}
	\centering
	\includegraphics[width=2.5in]{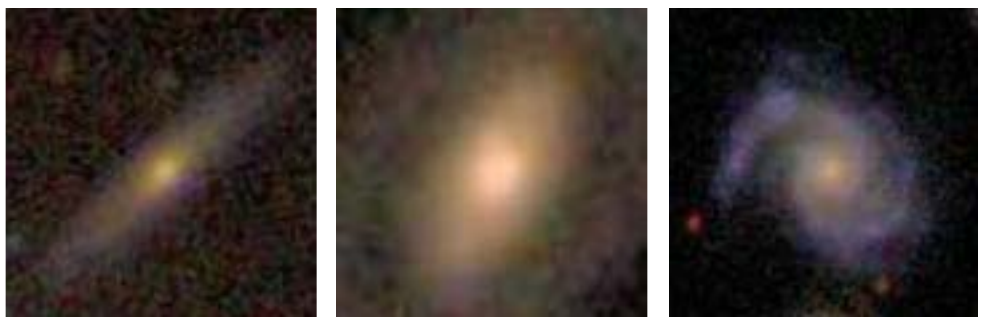}
	\caption{Images of the galaxy zoo}\label{fig:galaxy}
\end{figure}

We will first describe our datasets since they are different in bag numbers, class numbers, image contents, etc.
The selected datasets can be divided into four categories: (1)Object detection. Detect particular object in a image by the unique features of the object. The object often appears with complex background that could affect feature extraction. (2) Species recognition. A species may contains several classes. The task is to recognize the class of a species. The difficulty is that there exist exiguous difference between different classes of the same species.  (3) Galaxy discrimination. Discriminate the shape of galaxy in a image.  (4) Face identification. Identify the gender of a given face.
The detailed information of the datasets is introduced as follows.
\begin{itemize}
	\item Object detection: \textbf{Corel}\cite{duygulu2002object} is a famous natural scene image database which provides ten categories images, including architecture, bus, dinosaur, elephant, face, flower, food, horse, sky, snowberg. It was originally used for image annotation since every image contains several word annotations. But in our paper, we just use one main word for each image and make classification on the dataset.
	\textbf{Caltech} is another well-known and widely studied\cite{bosch2007image,zhang2006svm} image dataset for pattern recognition. The dataset is collected by the student from California Institute of Technology. We download six categories of images, including cars, motorcycles, airplanes, faces, leaves and backgrounds, from the website http://www.vision.caltech.edu. The third dataset \textbf{GRAZ02} is constructed by Andreas Opelt et.al in \cite{opelt2005object}. It contains four classes of images: bike, car, person and background(Environment without bike, car or person). Every image contains more complex contents than Corel and Caltech. For GRAZ02, two extra datasets are extracted, \textbf{Bike}(bike\&background) and \textbf{Car}(car\&background).
	Classification on the database is a challenging task due to clutter contents, illumination variation, intra-class variability, diversiform scales and poses.
	\item Species recognition: The image group of species, \textbf{butterfly}\cite{lazebnik2004semi}, is selected to make classification. The butterflies dataset consists of seven classes, admiral, black-swallowtail, machaon, monarch-closed, monarch-open, peacock, zebra. There exists slight difference between different kinds of butterflies, since they have similar shape, structure, and pose. So it is very hard to discriminate them, only their unique character(particular wing or stripe) may be helpful. 
	\item Galaxy discrimination: The \textbf{galaxy} zoo dataset consists of galaxy images with three kinds of shapes, edge-on, elliptical and spiral, which have been given in Figure \ref{fig:galaxy}. The database is collected by the Galaxy Zoo project(http://zoo1.galaxyzoo.org). Discriminate the shape of galaxy automatically is useful in astronomy since it can help scientists track the evolution of galaxies. 
	\item Face with Gender: The \textbf{FERET}(Facial Recognition Technology) database\cite{phillips2000feret} is established by the FERET program, whose primary task is to develop automatic face recognition technology that could be applied in assist security, intelligence and law enforcement personnel etc. We collect a part of the dataset and divide them into two classes, man and woman.
\end{itemize}

\begin{figure}	
\subfigure[Corel]{	
	\begin{minipage}{0.98\linewidth}
		\centering
			\includegraphics[width=0.5in]{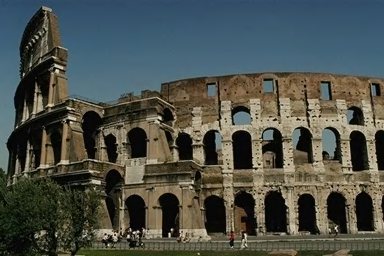}
			\includegraphics[width=0.5in]{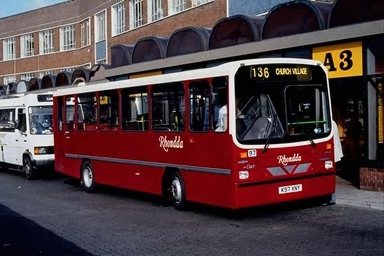}
			\includegraphics[width=0.5in]{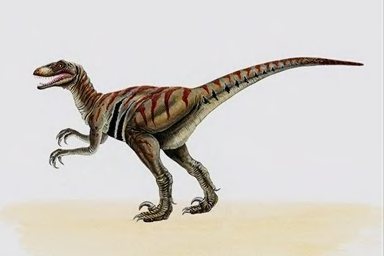}
			\includegraphics[width=0.5in]{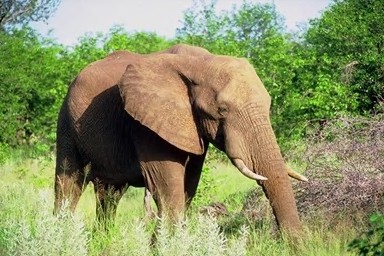}
			\includegraphics[width=0.5in]{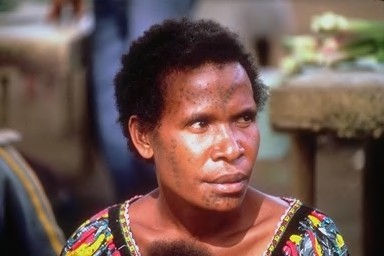}\\
			\includegraphics[width=0.5in]{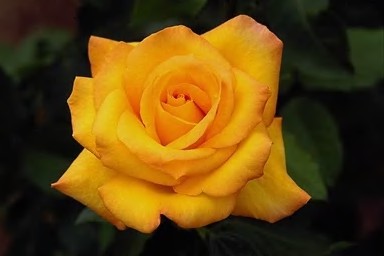}
			\includegraphics[width=0.5in]{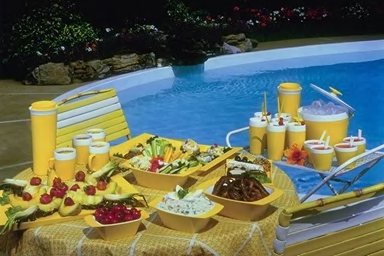}
			\includegraphics[width=0.5in]{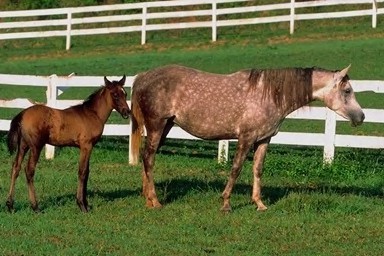}
			\includegraphics[width=0.5in]{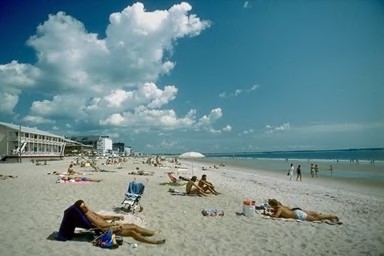}
			\includegraphics[width=0.5in]{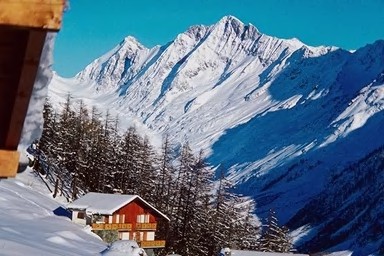}
			\par
			\vspace{0.1in}
	\end{minipage}
}
\subfigure[Caltech]{
	\begin{minipage}{0.98\linewidth}
		\centering
		\includegraphics[width=0.5in]{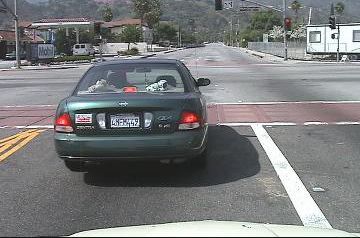}
		\includegraphics[width=0.5in]{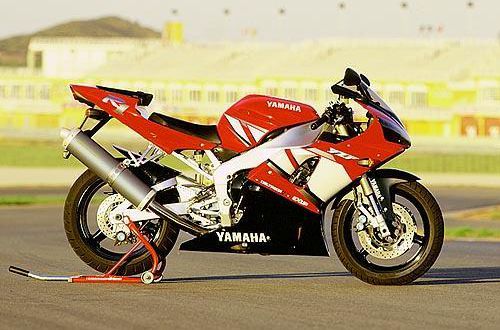}
		\includegraphics[width=0.5in]{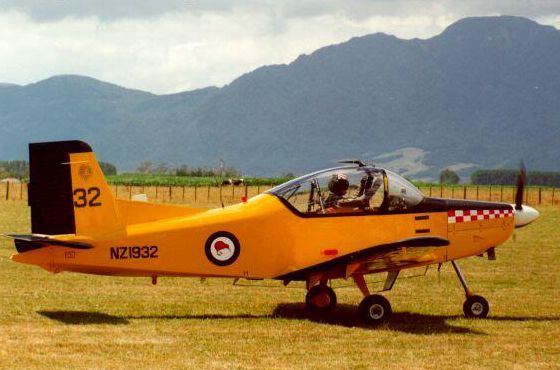}
		\includegraphics[width=0.5in]{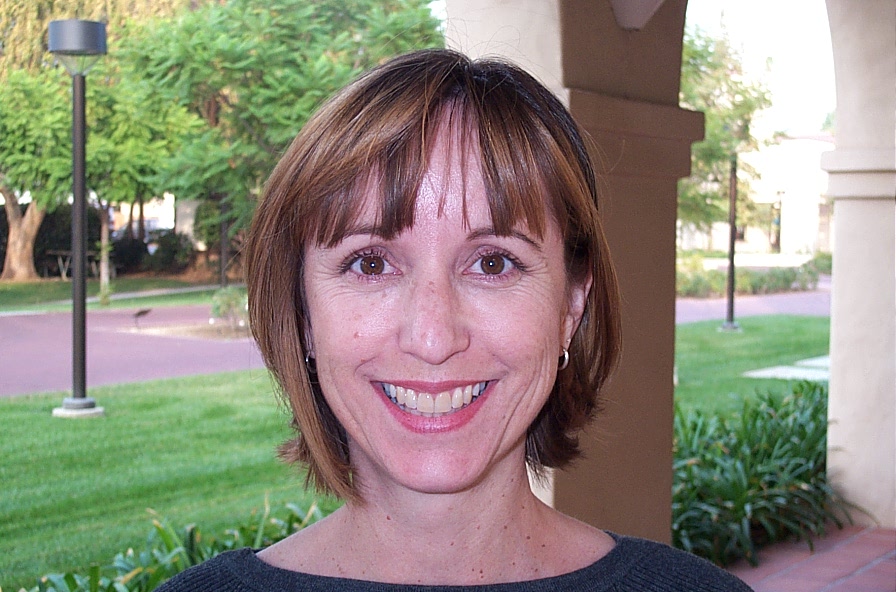}
		\includegraphics[width=0.5in]{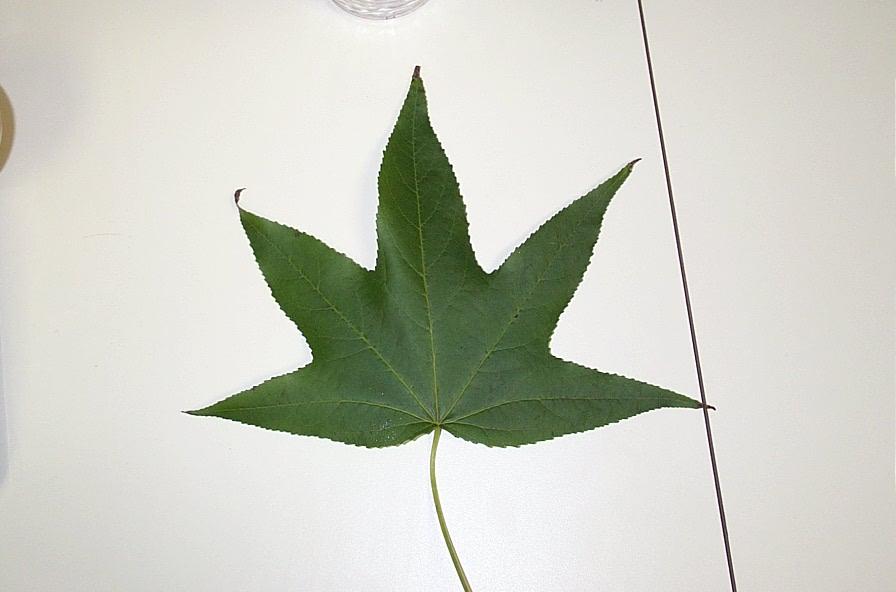}
		\includegraphics[width=0.5in]{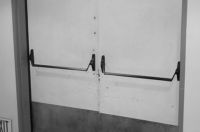}
		\par
		\vspace{0.1in}
	\end{minipage}
}
\subfigure[GRAZ02]{
	\begin{minipage}{0.98\linewidth}
		\centering
		\includegraphics[width=0.5in]{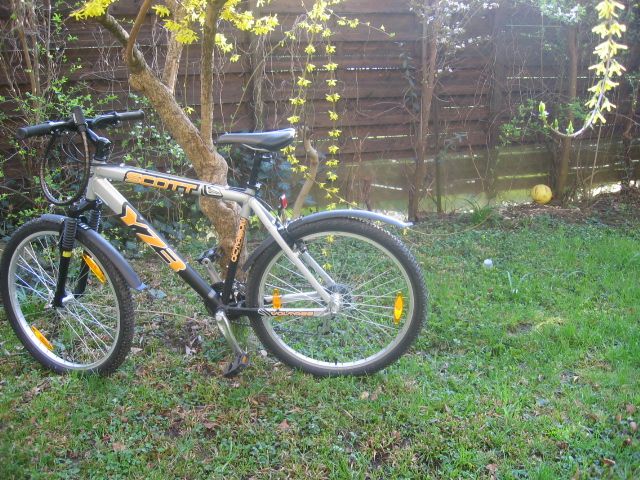}
		\includegraphics[width=0.5in]{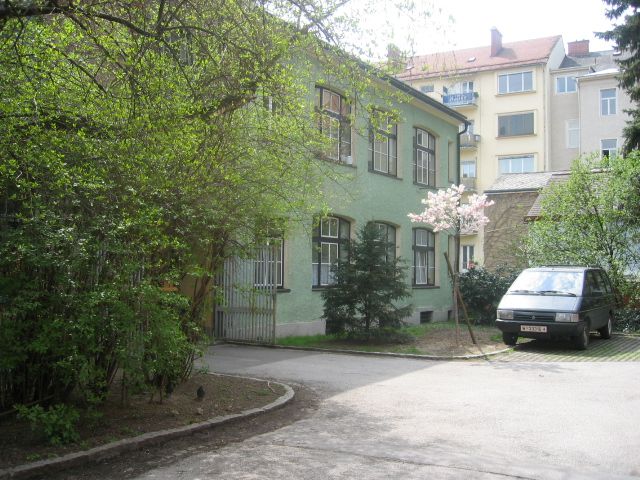}
		\includegraphics[width=0.5in]{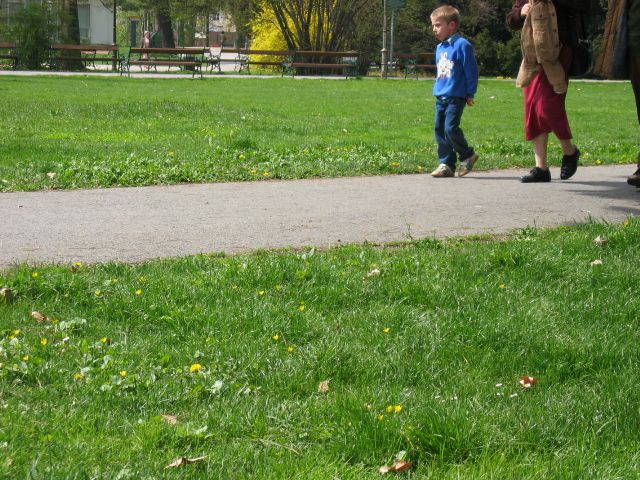}
		\includegraphics[width=0.5in]{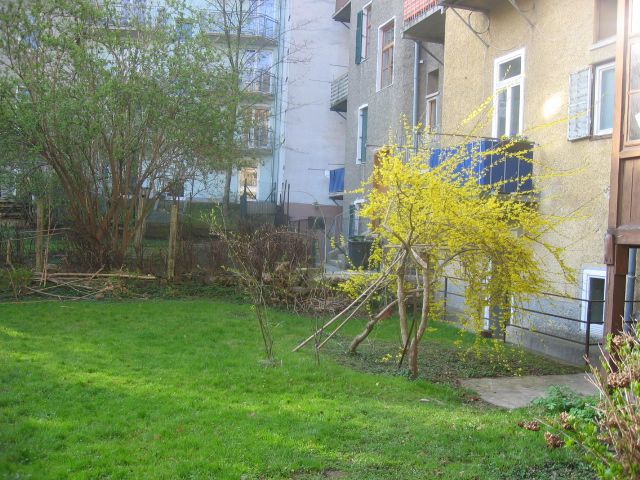}
		\par
		\vspace{0.1in}
	\end{minipage}
}
\caption{Images of Object Detection}\label{fig:object}
\end{figure}

\begin{figure}
	\centering
	\includegraphics[width=0.5in]{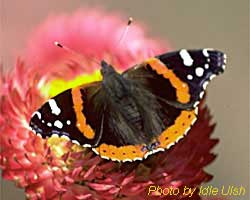}
	\includegraphics[width=0.5in]{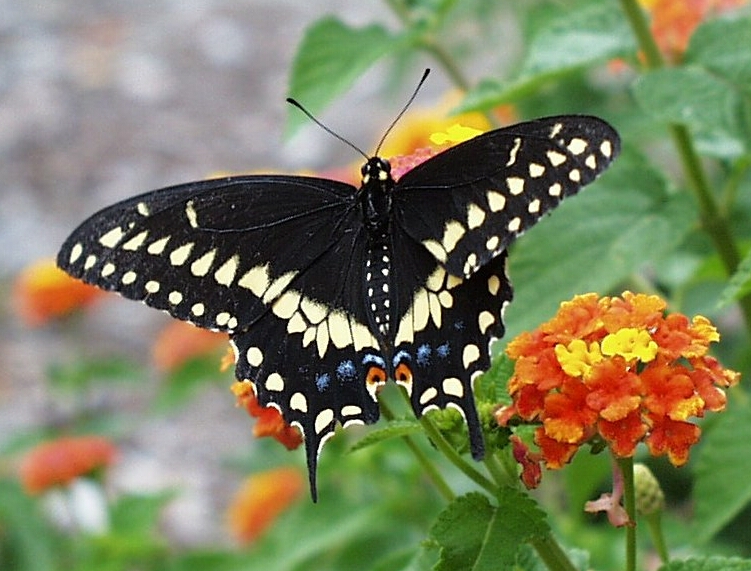}
	\includegraphics[width=0.5in]{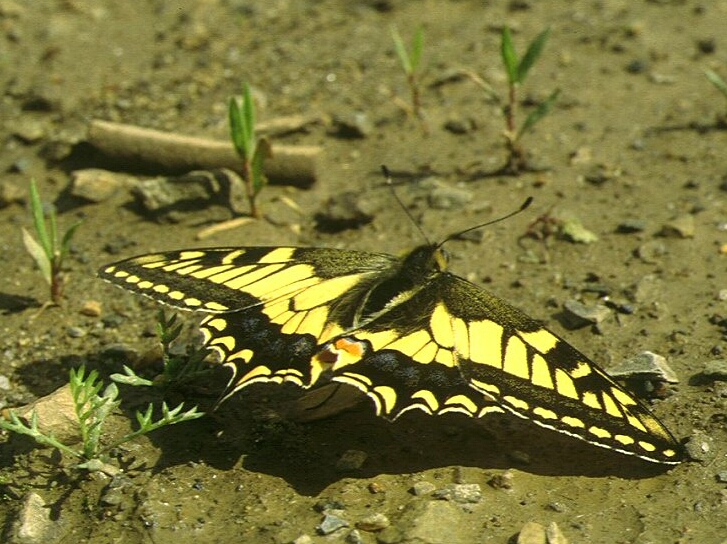}\\
	\par
	\vspace{0.1in}
	\includegraphics[width=0.5in]{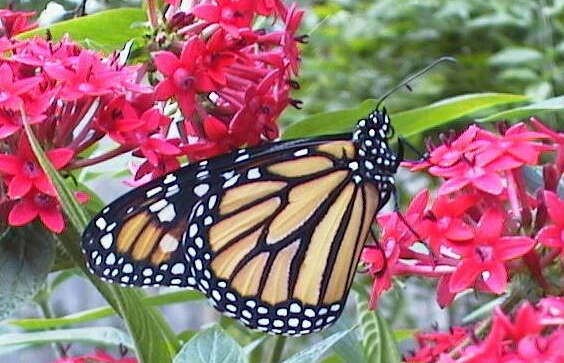}
	\includegraphics[width=0.5in]{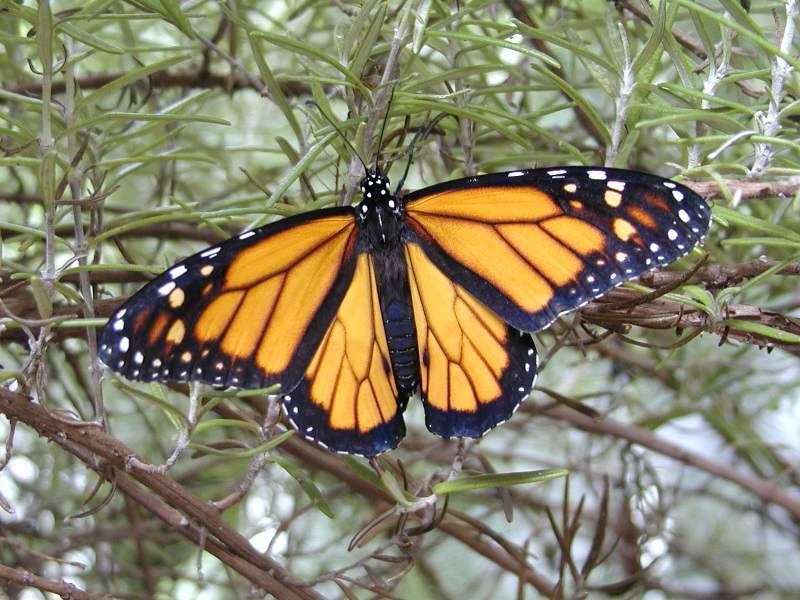}
	\includegraphics[width=0.5in]{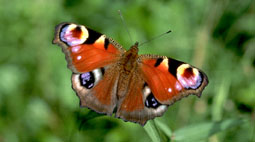}
	\includegraphics[width=0.5in]{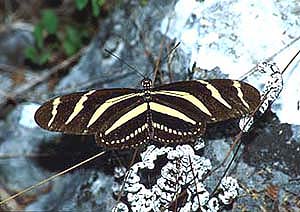}
	\caption{Images of the Butterfly}\label{fig:butterfly}
\end{figure}

\begin{figure}
		\begin{minipage}{0.98\linewidth}
			\centering
			\includegraphics[width=0.5in]{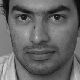}
			\includegraphics[width=0.5in]{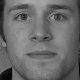}
			\includegraphics[width=0.5in]{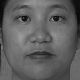}
			\includegraphics[width=0.5in]{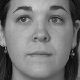}
			\par
			\vspace{0.1in}
		\end{minipage}
	\caption{Images of the FERET}
\end{figure}

\subsection{Feature extraction for images}
In this subsection, we will discuss what features will be extracted from images to construct different views of the datasets.

\subsubsection{HOG feature}
The HOG(Histogram of Oriented Gradient)\cite{dalal2005histograms} feature is presented based on the idea that local object appearance and shape can be well described by the distribution of local gradient or edge directions. It first divides image into smaller 'cells' and accumulates histogram of gradient or edge directions for each cell. The combination of these histograms represent the HOG feature of the whole image. However, we will not combine the histograms and then each image is a bag of cells, each cell can be expressed by a feature vector. In the following classification, each bag contains 9 cells.

\subsubsection{SIFT key points}
As a classical method in feature extraction, SIFT(Scale-invariant feature transformation) is first proposed by Lowe for object recognition and image matching\cite{lowe1999object,lowe2004distinctive}. It is a kind of local feature descriptor for images. It constructs scale space to get scale-invariant key points and endow orientation information with these points.
SIFT can find interest points at multiple scale to represent important regions of each image. Each key point is a 128-length numerical feature vector and each image is described by a bag of multiple such key points. It is worthy to point out that SIFT can extract different numbers of key points from different images, resulting that the idea of concatenating the vectors into a single feature vector is not tractable in traditional classification. The SIFT features are suitable to be used for multi-instance learning. SIFT has been widely applied in many applications\cite{felzenszwalb2010object,dollar2005behavior} since it is invariant to image scale and rotation. 

\subsubsection{Uniform patches with LBP}
\begin{figure}
	\centering
	\includegraphics[width=3.5in]{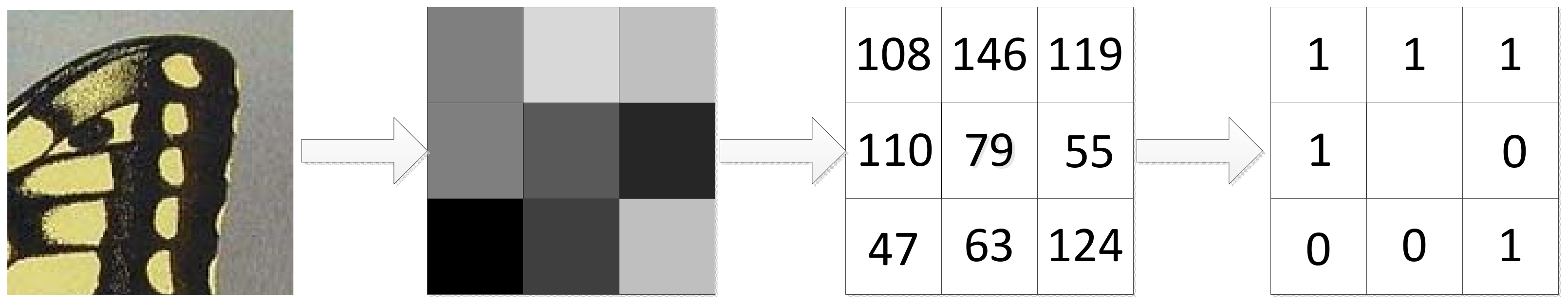}
	\caption{LBP}\label{fig:lbp}
\end{figure}

LBP(Local binary pattern) feature is a kind of operator to describe local textual property of image. Traditional LBP operator is defined in a $ 3 \times 3 $ block, the central pixel of which is used as the threshold. The other 8 neighbors are labeled as either 1 or 0 by comparing them with the threshold by turns. If the pixel value of neighbor is larger, it will be marked as 1, otherwise 0. Then a 8-bit string is obtained to represent the textual information of this local region. The 8-bit string will often be converted to numerical features for easily computation. Fig. \ref{fig:lbp} illustrates the process of extracting LBP features.
LBP has invariance of rotation and gray, making it be successfully applied in many applications, including image classification\cite{wang2009hog}, face detection\cite{zhang2007face}. In traditional, all the local LBP features are concatenated into a single feature vector to describe the whole image. 
An interesting concept 'visual dictionary' is proposed and used recently\cite{wen2009local} to describe image by partition it into smaller regions. Every region is a word of the dictionary and the image is represented by a bag of words. Similar as such method, every image will be divided into $ 4 \times 4 $ uniformly distributed patches, since multi-instance representation can be more powerful than single instance in improving the performance of classification, which we have mentioned in the Introduction. Every patch can be expressed by a single instance with length of 256 by extracting LBP features. Then each image is transformed as a bag, comprised of 16 instances.  

\subsection{Distance Comparison}
\begin{figure*}	
\subfigure[]{	
	\begin{minipage}{0.48\linewidth}
		\includegraphics[width=1.1in]{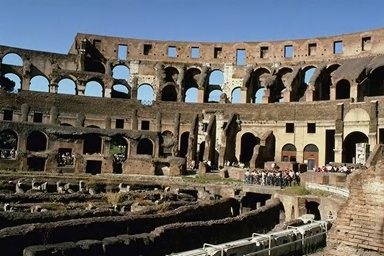}
		\begin{picture}(0,0)
		\put(-80,45){1}
		\end{picture}
		\hspace{-0.1in}
	\includegraphics[width=1.1in]{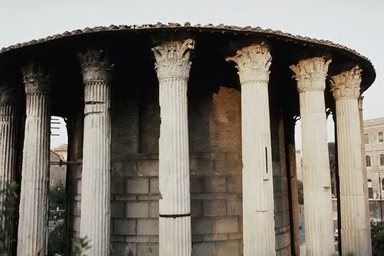}
			\begin{picture}(0,0)
			\put(-80,45){2}
			\end{picture}
			\hspace{-0.1in}
	\includegraphics[width=1.1in]{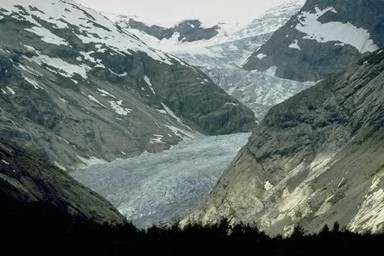}
			\begin{picture}(0,0)
			\put(-80,45){3}
			\end{picture}\\
	\includegraphics[width=1.1in]{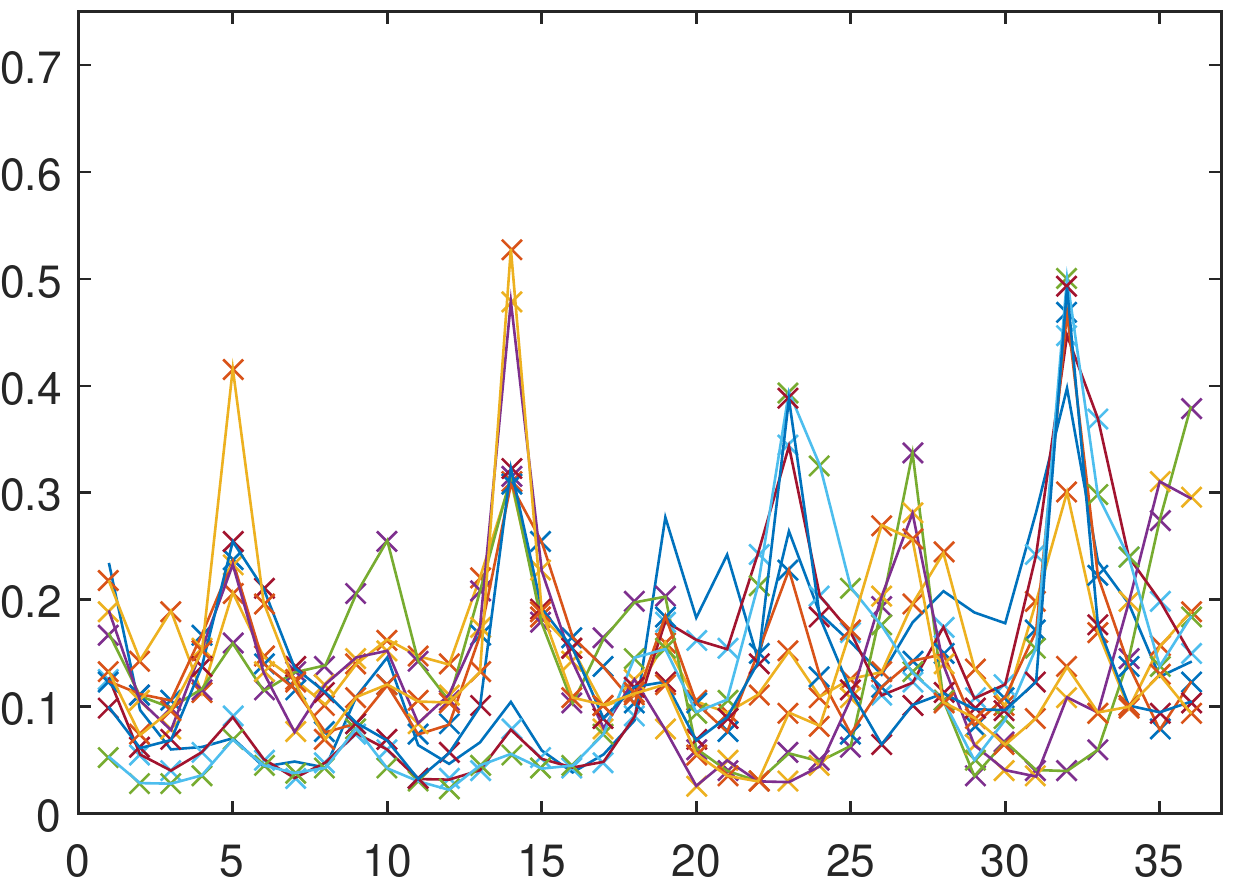}
	\includegraphics[width=1.1in]{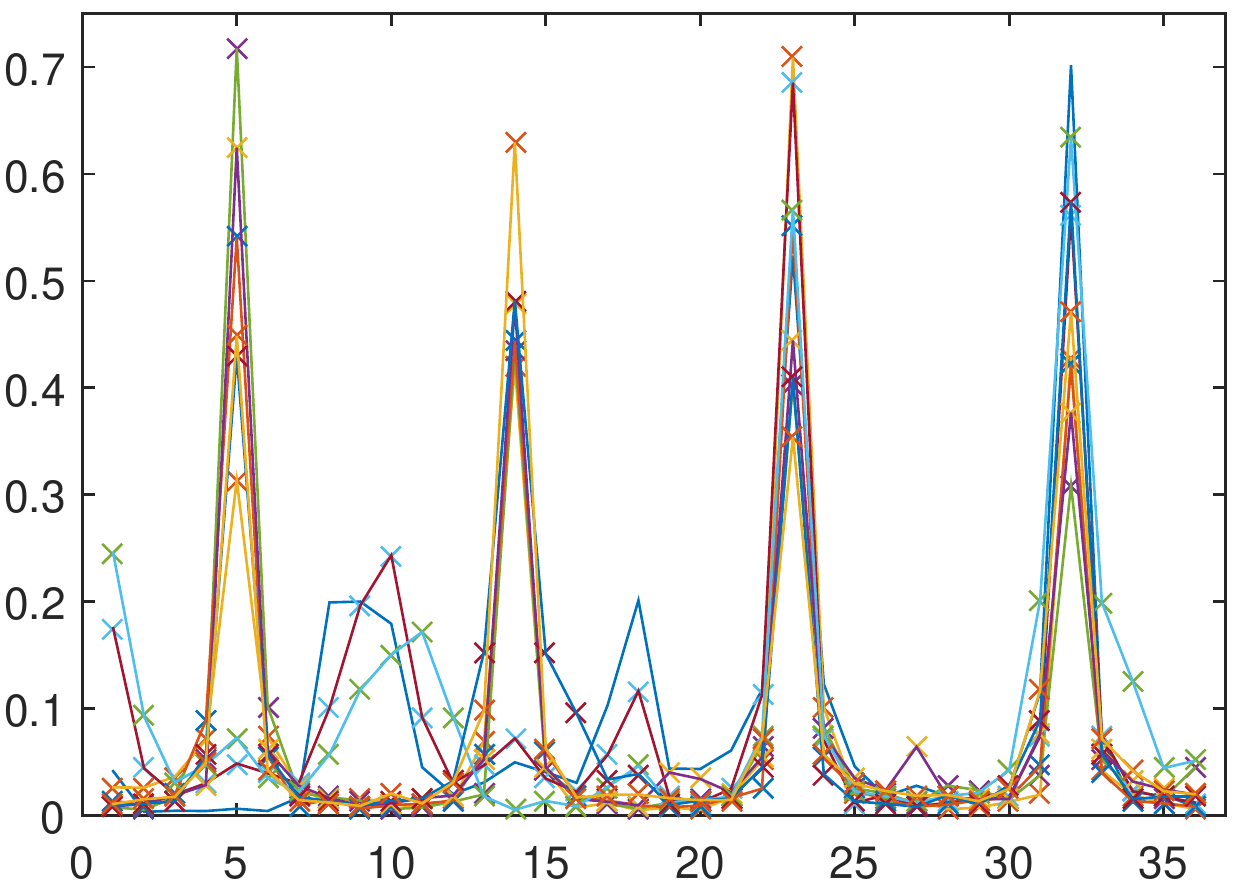}
	\includegraphics[width=1.1in]{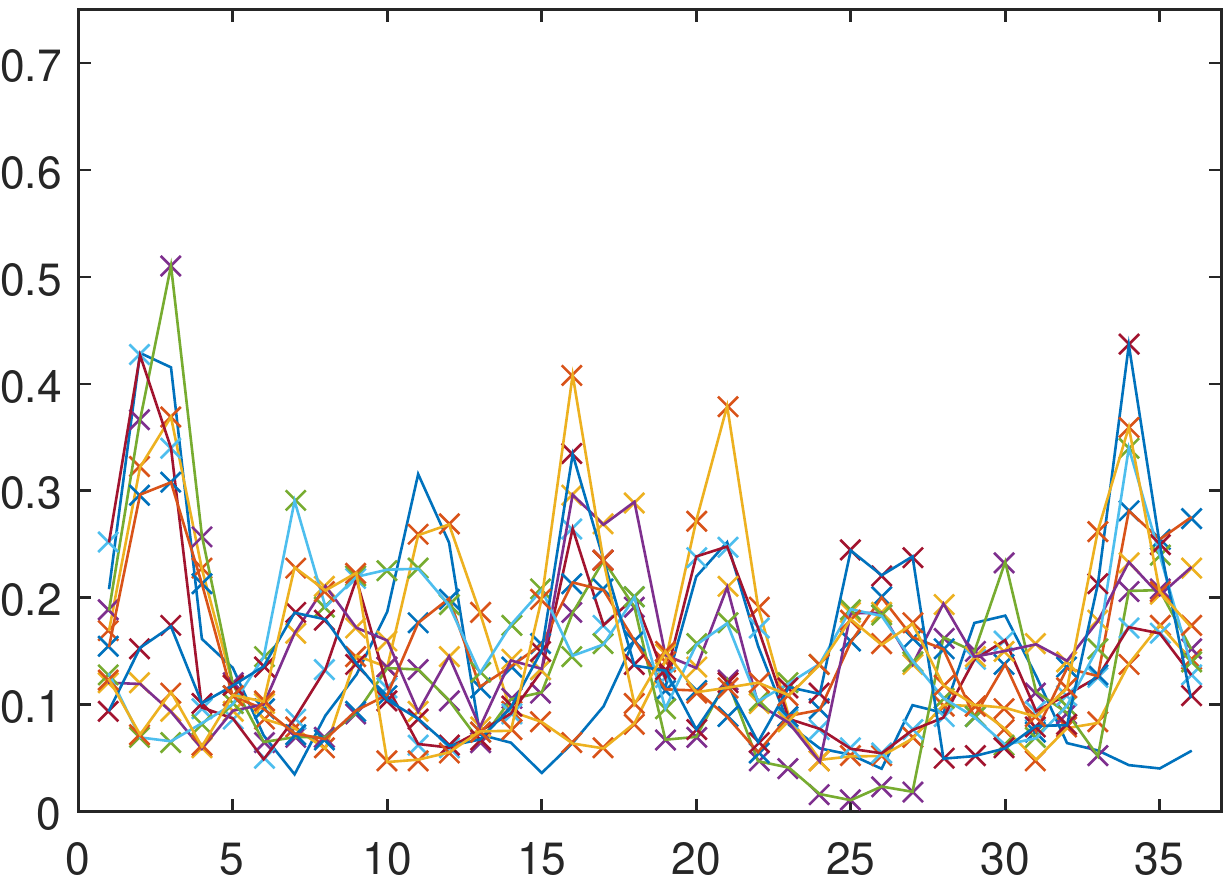}\\
	\includegraphics[width=1.1in]{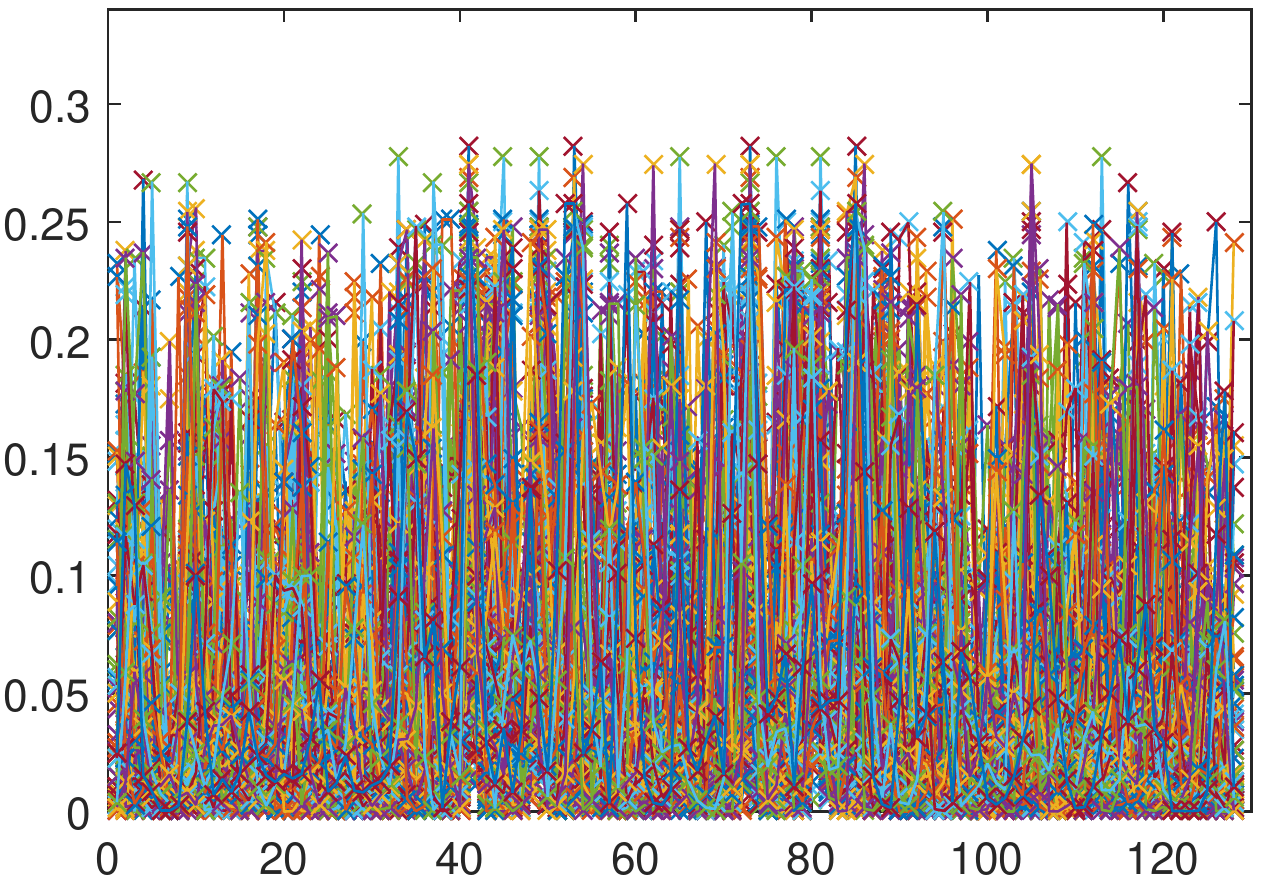}
	\includegraphics[width=1.1in]{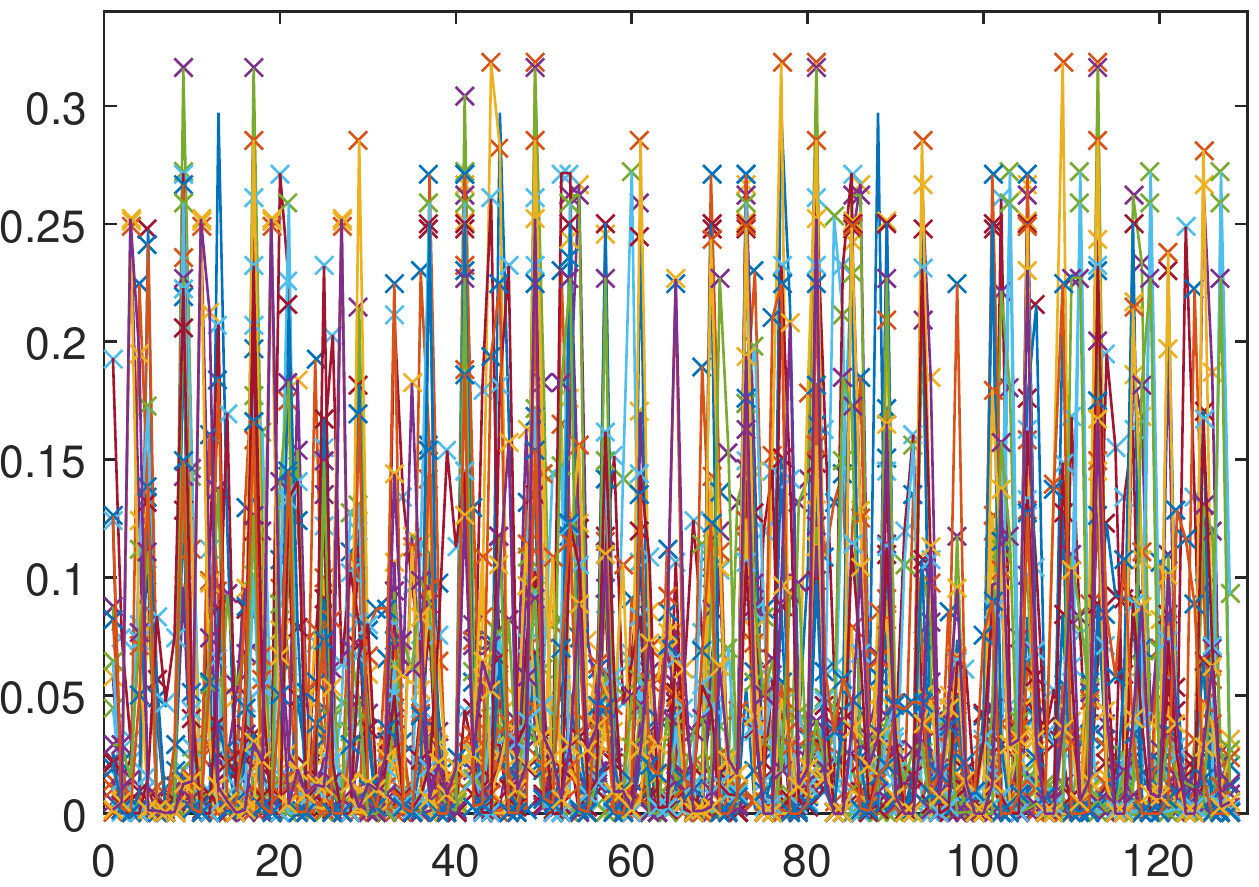}
	\includegraphics[width=1.1in]{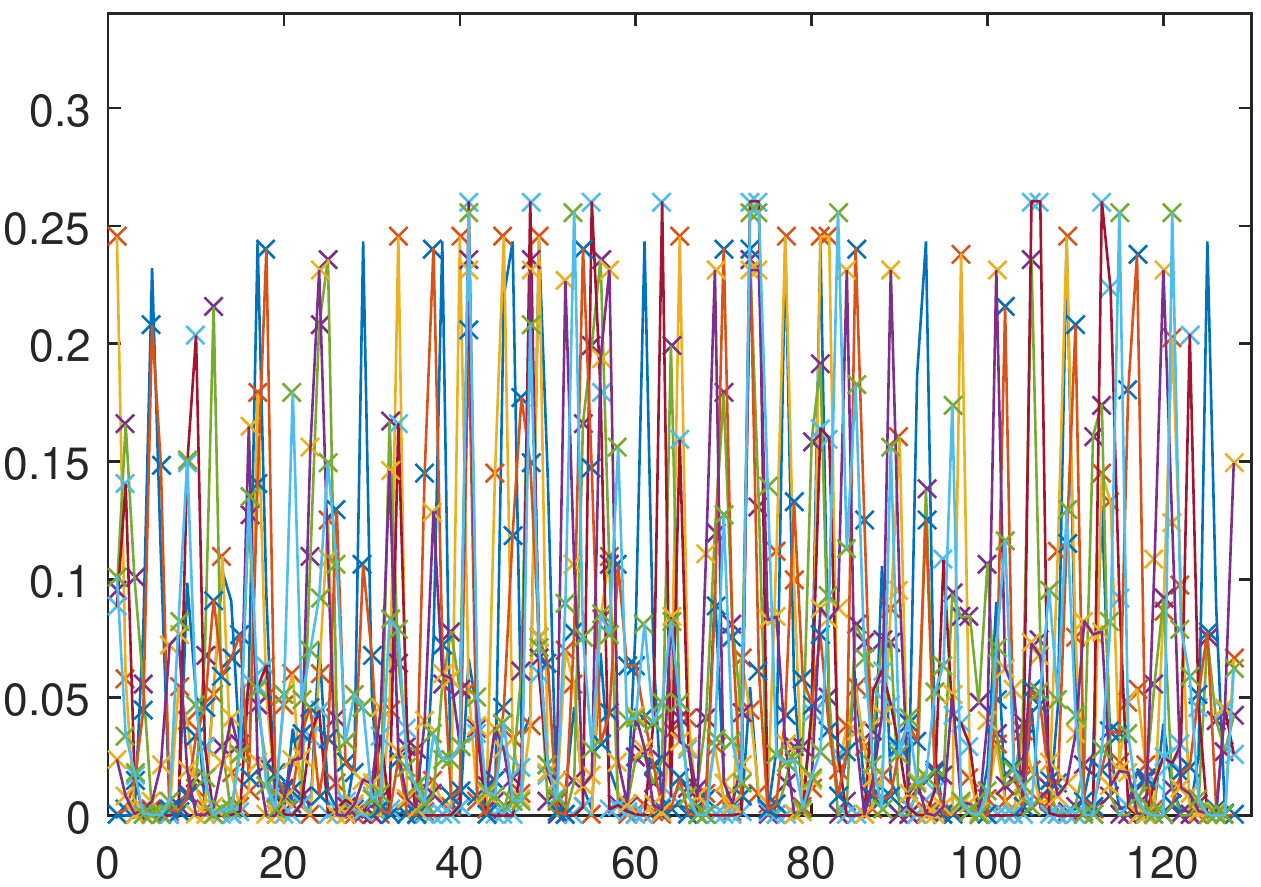}\\
	\includegraphics[width=1.1in]{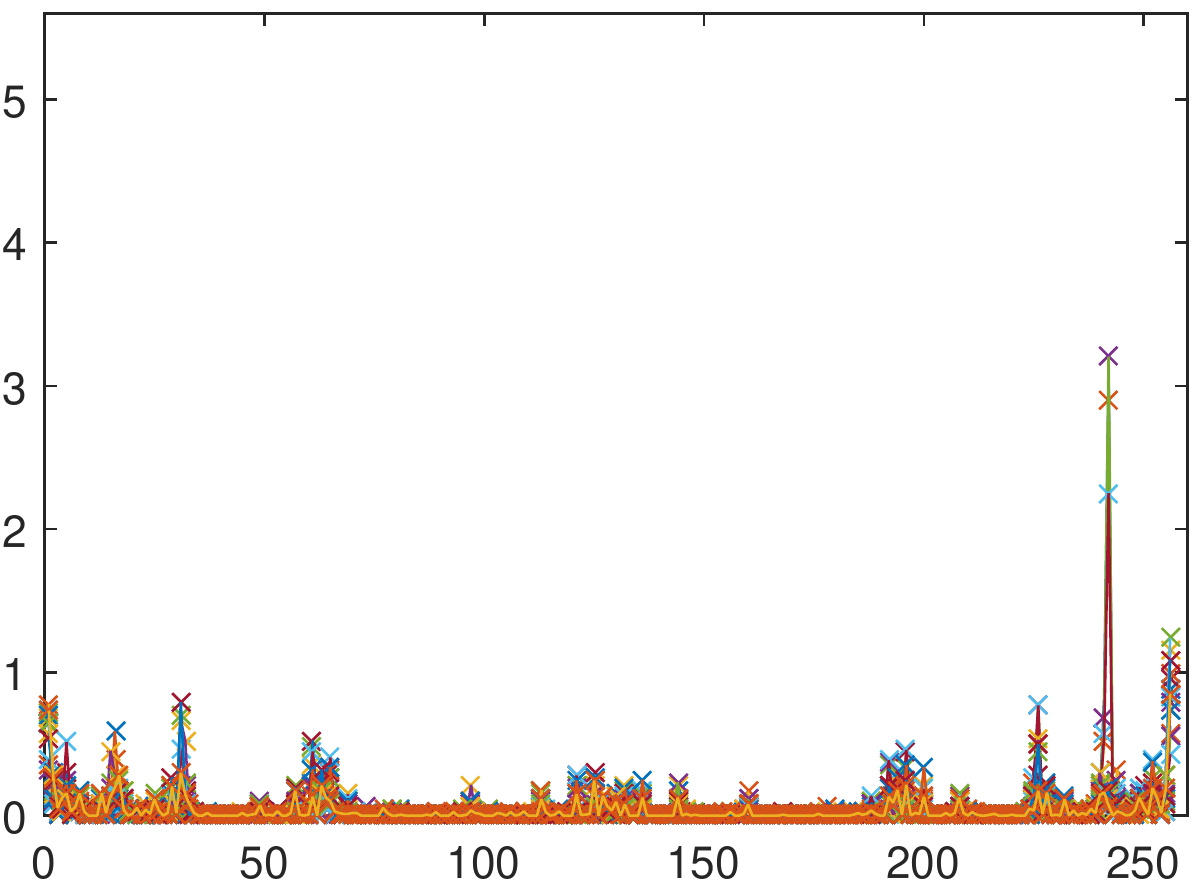}
	\includegraphics[width=1.1in]{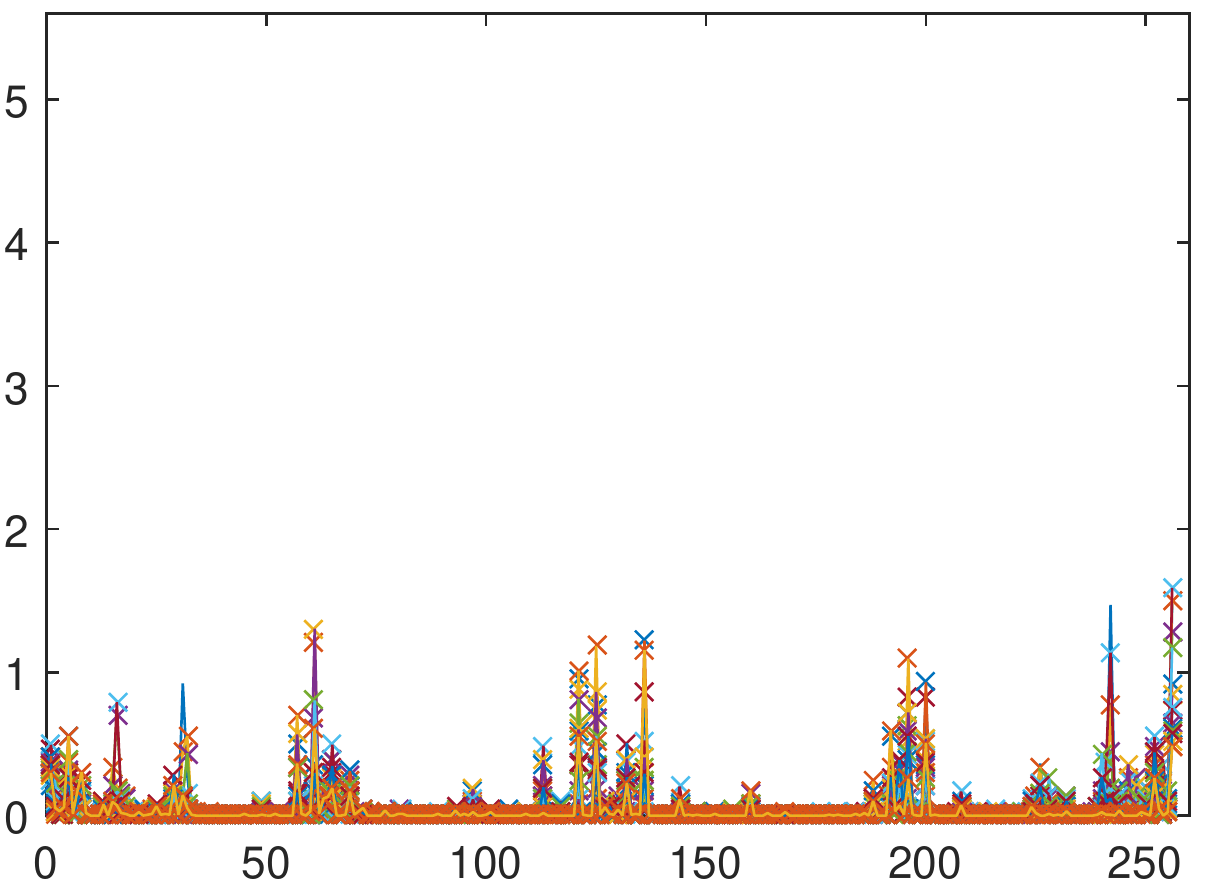}
	\includegraphics[width=1.1in]{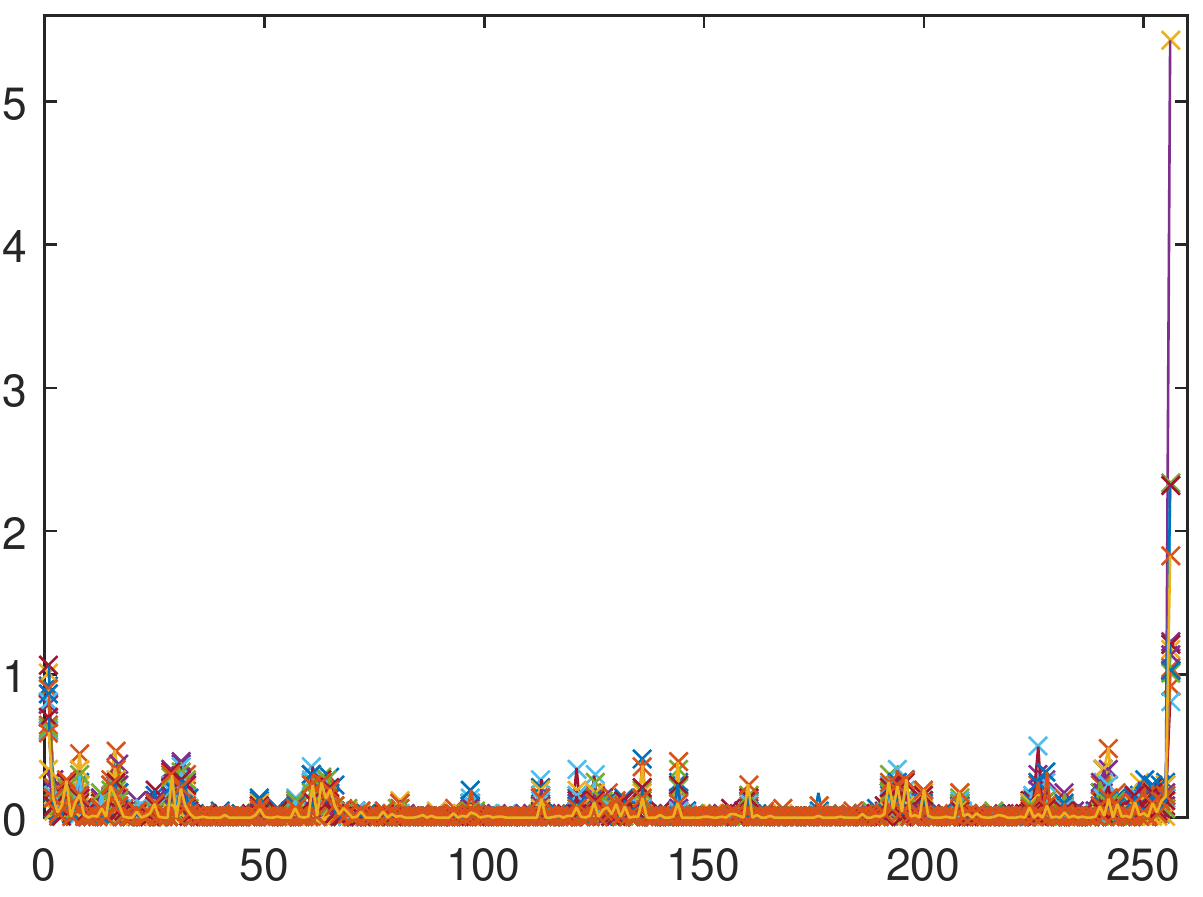}
			\par
			\vspace{0.1in}
\end{minipage}
}
\subfigure[]{
	\begin{minipage}{0.55\linewidth}
		\scriptsize
		\renewcommand{\arraystretch}{2}
        \begin{tabular}{lc|c|c|}
				& \multicolumn{3}{c}{$ D_{am} $} \\
				\cline{2-4}
				\multirow{3}{0.5cm}{HOG} &\multicolumn{1}{|c|}{0} & \underline{0.49} & \underline{0.41} \\\cline{2-4}
				&\multicolumn{1}{|c|}{\underline{0.49}} & 0 & 1.14 \\\cline{2-4}
				&\multicolumn{1}{|c|}{\underline{0.41}} & 1.14 & 0 \\\cline{2-4}
        \end{tabular}		
		\begin{tabular}{|c|c|c|}
			\multicolumn{3}{c}{$ D_{min} $} \\
			\hline
			0 & \underline{0.29} & \underline{0.26}  \\\hline
			\underline{0.29} & 0 & 1.03  \\\hline
			\underline{0.26} & 1.03 & 0  \\\hline
		\end{tabular}
		\begin{tabular}{|c|c|c|}
			\multicolumn{3}{c}{$ D_{ave} $} \\
			\hline
			0.38 & \underline{0.81} & \underline{0.61}  \\\hline
			\underline{0.81} & 0.37 & 1.30  \\\hline
			\underline{0.61} & 1.30 & 0.34  \\\hline
		\end{tabular}\\
	\par
	\vspace{0.05in}	
		\begin{tabular}{l|c|c|c|}
			\cline{2-4}
		\multirow{3}{0.5cm}{SIFT} &0 & \textbf{0.66} & 0.78\\\cline{2-4}
			&\textbf{0.66} & 0 & 0.72 \\\cline{2-4}
			&0.78 & 0.72 & 0 \\\cline{2-4}
		\end{tabular}
		\begin{tabular}{|c|c|c|}\hline
			0 & \textbf{0.30} & 0.48  \\\hline
			\textbf{0.30} & 0 &  0.45  \\\hline
			0.48 & 0.45 & 0  \\\hline
		\end{tabular}
		\begin{tabular}{|c|c|c|}\hline
			0.94 & \underline{1.08} & \underline{1.07} \\\hline
			\underline{1.08} & 0.95 & \underline{1.07} \\\hline
			\underline{1.07} & \underline{1.07} & 0.94  \\\hline
		\end{tabular}
	\par
	\vspace{0.05in}
	\begin{tabular}{l|c|c|c|}\cline{2-4}
	\multirow{3}{0.5cm}{LBP} &	0 & \textbf{1.69} & 1.96 \\\cline{2-4}
		&\textbf{1.69} & 0 & 2.21 \\\cline{2-4}
		&1.96 & 2.21 & 0 \\\cline{2-4}
	\end{tabular}
	\begin{tabular}{|c|c|c|}\hline
		0 & \underline{0.49} & \underline{0.38}  \\\hline
		\underline{0.49} & 0 &  0.67  \\\hline
		\underline{0.38} & 0.67 & 0  \\\hline
	\end{tabular}
	\begin{tabular}{|c|c|c|}\hline
		3.86 & \underline{4.41} & 4.57 \\\hline
		\underline{4.41} & 2.82 & \underline{4.37} \\\hline
		4.57 & \underline{4.37} & 3.07  \\\hline
	\end{tabular}
		\par
		\vspace{0.1in}	
	\end{minipage}
}
	\caption{A toy example to compare three distance functions $ D_{am}, D_{min} $ and $ D_{ave} $. (a) The 2, 3, 4 line of images corresponds to HOG, SIFT, and LBP respectively. Each curve in a subfigure denotes a instance. The y axis denotes the numerical value of the components of instance. (b) The computed distances in three kinds of features. The minimum in each sub-table is in boldface. The numbers with underline can misguide the judge of distance function.}
	\label{fig:distcmp}
\end{figure*}

In the Section \ref{sec:dist}, we have claimed that our novel distance function $ D_{am} $, designed for bags, is prior to the previous $ D_{ave} $ and $ D_{min} $. Next we will make numerical experiments to verify the judgement. First, a toy example is given in the Figure \ref{fig:distcmp} to shown the difference among the three distance functions. Three images from the \textbf{Corel} dataset were selected. The former two belong to the class of \textit{architecture} and the third belongs to \textit{snowberg}. The features of HOG, SIFT, LBP are extracted and depicted in the Figure \ref{fig:distcmp}(a). The distances were computed in the Figure \ref{fig:distcmp}(b). In fact, the three images are similar in structure, color and luminance, which gives a challenge to distance function.
In SIFT and LBP feature, $ D_{am}(I_1,I_2) $ is smaller than $ D_{am}(I_1,I_3) $ and $ D_{am}(I_2,I_3) $, implying that $ I_1 $ and $ I_2 $ belong to the same class. So $ D_{am} $ made right judges in SIFT and LBP feature, better than $ D_{ave} $ and $ D_{min} $.

To further validate the efficiency and advantage of $ D_{am} $, $1$NN classification was implemented with the three distance functions respectively. Euclidean distance was used to compute the distance between instances.
Three datasets, \textbf{Car}, \textbf{butterfly} and \textbf{Corel}, were selected to make comparison. Three-fold cross validation was applied in classification. Accuracy and standard deviation are shown in the Table \ref{tab:dist1}. It can be clearly seen that $1$NN with $ D_{am} $ obtains the best performance on most of the features of the three datasets. 

\begin{table*}[htbp]
	\small
	\centering	
	\renewcommand{\arraystretch}{1.1}
	\caption{1-NN classification accuracy of different distances on different features}\label{tab:dist1}
	\begin{tabular}{c|l|ccc|c}
		\hline
\multirow{2}{*}{Datasets} & \multirow{2}{*}{Distance}  & \multicolumn{3}{c|}{Feature} & Average \\
\cline{3-5}
& & HOG & SIFT & LBP & \\
\hline
\multirow{3}*{\tabincell{c}{Car\\(600\&2)}} & $ D_{ave} $ & 56.67$\pm$2.25  & 51.83$\pm$1.04 & 47.50$\pm$6.56 & 52.00 \\
& $ D_{min} $ & 60.67$\pm$2.52 &  59.33$\pm$2.36 & 53.83$\pm$1.61 & 57.94 \\
& $ D_{am} $ & \textbf{64.00$\pm$3.46} & \textbf{64.50$\pm$2.00 }& \textbf{59.17$\pm$1.89} & 62.56 \\
\hline
\multirow{3}{*}{\tabincell{c}{Butterfly\\(619\&7)}} & $ D_{ave} $ & 17.28$\pm$3.64  &  23.42$\pm$3.16 & 21.16$\pm$1.97 & 20.62 \\
& $ D_{min} $ & 40.55$\pm$2.75 &  \textbf{84.33$\pm$0.98} & 26.66$\pm$1.31 & 50.51 \\
& $ D_{am} $ & \textbf{52.66$\pm$1.89} & 81.74$\pm$1.49 & \textbf{28.12$\pm$4.73} & 54.17 \\
\hline
\multirow{3}{*}{\tabincell{c}{Corel\\(1000\&10)}} & $ D_{ave} $ & 15.20$\pm$4.53  &  13.60$\pm$0.44 & 36.50$\pm$3.45 & 21.77 \\
& $ D_{min} $ & 44.50$\pm$1.28 &  32.90$\pm$1.08 & 39.80$\pm$3.05 & 39.07 \\
& $ D_{am} $ & \textbf{50.70$\pm$3.47} & \textbf{46.30$\pm$3.06} & \textbf{58.00$\pm$1.32} & 51.67 \\
\hline	
	\end{tabular}
\end{table*}

\subsection{Performance evaluation}
In the subsection, we will evaluate our model in different sides, including classification ability , robustness to parameters and sensitivity to instance number of bags.
\subsubsection{Image classification}
Comparison of the classification performance will be made eight datasets, the information of which is given in the Table \ref{tab:error3}. Three-fold cross-validation was utilized to get the average classification accuracy. The experiments had been divided into two parts: single view and multi-view. 
In single view classification, experiments on the three features HOG, SIFT and LBP were conducted independently. $ k $NN classification with Euclidean distance and metric learning(Algorithm 1) were both implemented. In metric learning, the penalty parameter $ \lambda $ and the learning rate $ \eta $ were both empirically set to be 0.1 and the number of iteration $ R $ was set to be 3.
It is obvious that single view metric learning can always improve the performance of $ k $NN classification with Euclidean distance. SVML is effective in obtaining better metric for images with multi-instance features. 
For multi-view classification, the three features can be combined into four groups: HOG+SIFT(H\&S), HOG+LBP(H\&L), SIFT+LBP(S\&L), HOG+SIFT+LBP(H\&S\&L). The experiments on these four multi-view conditions were made individually.
The penalty parameters $ \lambda $ and $ \mu $ are both selected from the set $ \{0.01, 0.1, 1\} $ and the combination of $ \eta_1, \eta_2 $ is chosen from the set $ \{(0.01,0.01), (0.05,0.05), (0.1,0.1)\} $. The number of iteration $ \tau, R $ is set to be 1 and 3 respectively.
The classification results are shown in the Table \ref{tab:error3}. Experiments with the feature group H\&S\&L perform best on 5 datasets, better than single view classification, despite metric learning or not. Classification with more views often get higher accuracy, which indicates that our method can extract useful information from all the views and assemble them effectually, based on the inference that different features are complementary to each other. Further, six images from three datasets are displayed in the Table \ref{tab:nearimg} with their corresponding nearest images under different views. It implies that our method can truly find a data-dependent metric to make similar images closer and improve the classification performance.

\begin{table*}[htbp]
\small
\centering	
\renewcommand{\arraystretch}{1.3}
\caption{Classification accuracy on single view and multi-view}\label{tab:error3}
\begin{tabular}{l|lccc|cccc}
\hline
\multirow{2}{*}{Datasets} & \multicolumn{4}{c|}{Single view} & \multicolumn{4}{c}{Multi-view} \\
\cline{2-9}
       &  & HOG & SIFT & LBP &  H\&S  & H\&L & S\&L & H\&S\&L \\
\hline
\multirow{2}{*}{\tabincell{c}{Corel\\(300\&10)}} & Eucl.& 43.00$\pm$2.65 & 42.67$\pm$4.16 & 54.00$\pm$2.00 & \multirow{2}{*}{52.33$\pm$6.11} & \multirow{2}{*}{55.33$\pm$2.52} & \multirow{2}{*}{57.67$\pm$6.43} &  \multirow{2}{*}{\textbf{60.00$\pm$1.73}} \\
& ML& 44.00$\pm$1.00 & 42.67$\pm$4.16 & 59.33$\pm$5.77 &  &  &  &\\
\hline
\multirow{2}{*}{\tabincell{c}{Caltech\\(300\&6)}} & Eucl.& 80.33$\pm$3.51 & 66.67$\pm$1.53 & 78.67$\pm$3.21 & \multirow{2}{*}{86.00$\pm$2.65}  & \multirow{2}{*}{85.00$\pm$6.08} & \multirow{2}{*}{77.33$\pm$1.53} & \multirow{2}{*}{\textbf{87.00$\pm$4.58}} \\
& ML& 80.33$\pm$2.52 & 67.67$\pm$1.53 & 78.33$\pm$3.06 & & & &\\
\hline
\multirow{2}{*}{\tabincell{c}{GRAZ02\\(300\&4)}} & Eucl.& 37.33$\pm$8.96 & 38.67$\pm$2.89 & 36.67$\pm$3.79 & \multirow{2}{*}{36.67$\pm$4.73} & \multirow{2}{*}{35.00$\pm$4.58} &\multirow{2}{*}{37.33$\pm$3.79}& \multirow{2}{*}{37.33$\pm$4.73} \\
& ML& 38.67$\pm$8.50 & \textbf{40.00$\pm$3.61} & 37.33$\pm$4.16 & & & &\\
\hline
\multirow{2}{*}{\tabincell{c}{Bike\\(300\&2)}} & Eucl.& \textbf{58.51$\pm$1.00} & 53.57$\pm$6.79 & 56.52$\pm$3.54 & \multirow{2}{*}{55.04$\pm$3.18} &    \multirow{2}{*}{58.04$\pm$4.92} & \multirow{2}{*}{58.04$\pm$4.92} &\multirow{2}{*}{58.48$\pm$7.74} \\
& ML& 58.51$\pm$1.00 & 53.60$\pm$8.52 & 57.55$\pm$7.90 & & & &\\
\hline
\multirow{2}{*}{\tabincell{c}{Car\\(300\&2)}} & Eucl.& 60.49$\pm$2.53 & 58.11$\pm$11.3 & 54.98$\pm$8.58 & \multirow{2}{*}{\textbf{68.54$\pm$4.02}} & \multirow{2}{*}{62.97$\pm$2.88}& \multirow{2}{*}{67.01$\pm$2.38}& \multirow{2}{*}{64.53$\pm$2.51} \\
& ML& 59.98$\pm$4.40 & 58.10$\pm$9.84 & 56.97$\pm$4.76 & & & &\\
\hline
\multirow{2}{*}{\tabincell{c}{Butterfly\\(280\&7)}} & Eucl.& 47.51$\pm$2.93 & 71.43$\pm$1.34& 22.85$\pm$1.50& \multirow{2}{*}{73.22$\pm$0.93} & \multirow{2}{*}{43.25$\pm$8.13} & \multirow{2}{*}{67.14$\pm$4.15} & \multirow{2}{*}{\textbf{73.57$\pm$2.29}} \\
 & ML& 48.22$\pm$2.02 & 71.79$\pm$0.44 & 24.64$\pm$0.15 &  & & &\\
 \hline
\multirow{2}{*}{\tabincell{c}{Galaxy\\(210\&3)}} & Eucl.& 78.10$\pm$0.82 & 62.38$\pm$1.65 & 81.90$\pm$8.37 & \multirow{2}{*}{81.90$\pm$4.12}  & \multirow{2}{*}{83.33$\pm$3.60} & \multirow{2}{*}{84.29$\pm$5.71} & \multirow{2}{*}{\textbf{85.71$\pm$3.78}} \\
 & ML& 78.10$\pm$0.82 & 62.86$\pm$1.43 & 82.38$\pm$7.87 & & & & \\
 \hline
\multirow{2}{*}{\tabincell{c}{FERET\\(150\&2)}} & Eucl.& 82.67$\pm$2.31 & 76.00$\pm$7.21 & 71.33$\pm$5.03& \multirow{2}{*}{84.00$\pm$8.72} & \multirow{2}{*}{81.33$\pm$1.15} & \multirow{2}{*}{78.00$\pm$4.00} & \multirow{2}{*}{\textbf{84.00$\pm$2.00}}	\\
 & ML& 82.67$\pm$2.31 & 76.00$\pm$5.29 & 76.00$\pm$2.00 & & & & \\
 \hline	
\end{tabular}
\end{table*}

\begin{table*}[htbp]
	\centering	
	\renewcommand{\arraystretch}{1.2}
	\caption{Nearest Image}\label{tab:nearimg}
	\begin{tabular}{cccccccc}		
		Image & HOG & SIFT & LBP & H\&S & H\&L & S\&L & H\&S\&L \\
		\includegraphics[width=0.7in]{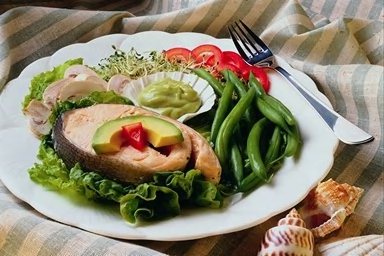} & \includegraphics[width=0.7in]{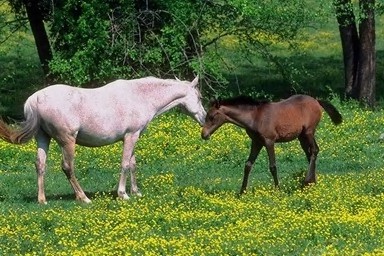} & \includegraphics[height=0.5in]{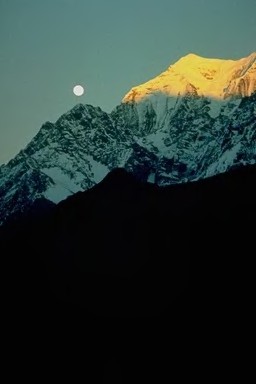} & \includegraphics[width=0.7in]{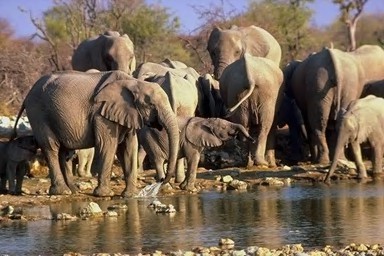} &
		\includegraphics[width=0.7in]{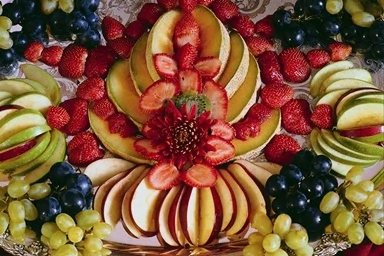} & \includegraphics[width=0.7in]{NearestImg/corel216.jpg} & \includegraphics[height=0.5in]{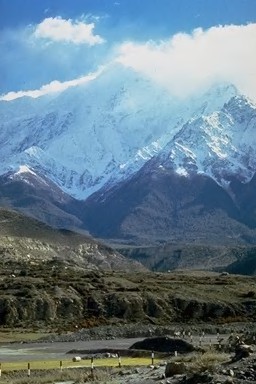}& \includegraphics[width=0.7in]{NearestImg/corel187.jpg}\\
		\includegraphics[width=0.7in]{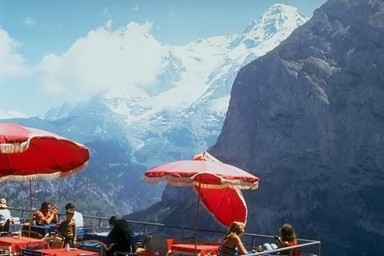} & \includegraphics[height=0.5in]{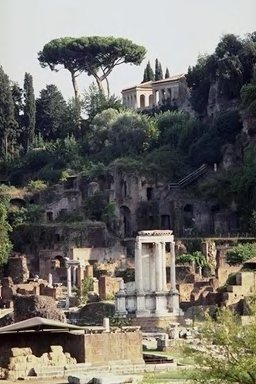} & \includegraphics[width=0.7in]{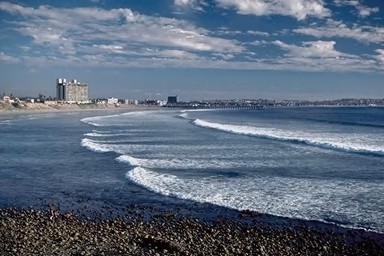} & \includegraphics[width=0.7in]{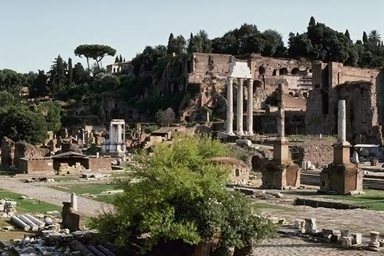} &
		\includegraphics[width=0.7in]{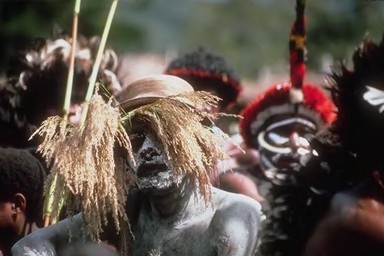} & \includegraphics[height=0.5in]{NearestImg/corel10.jpg} & \includegraphics[width=0.7in]{NearestImg/corel254.jpg}& \includegraphics[width=0.7in]{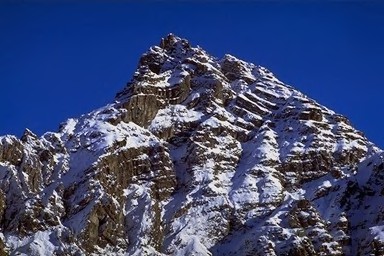}\\
		\includegraphics[width=0.7in]{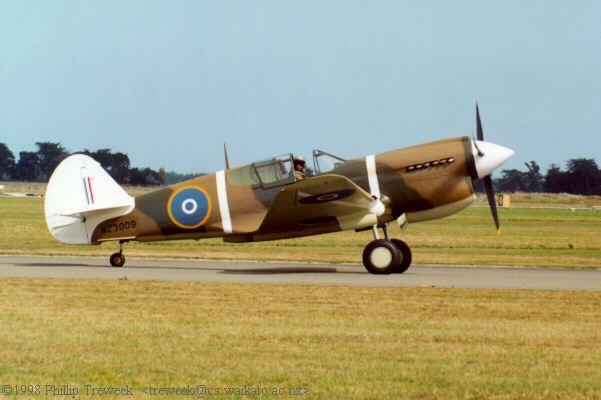} & \includegraphics[width=0.7in]{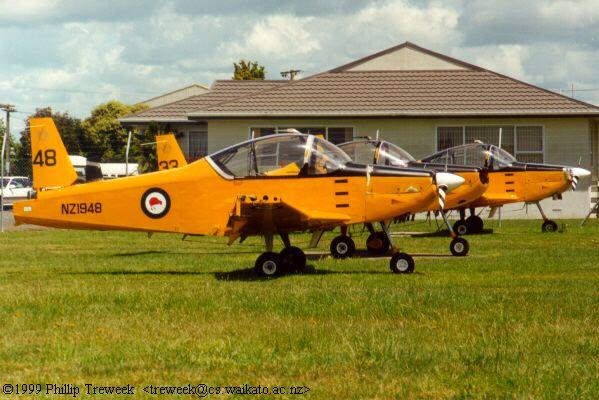} & \includegraphics[width=0.7in]{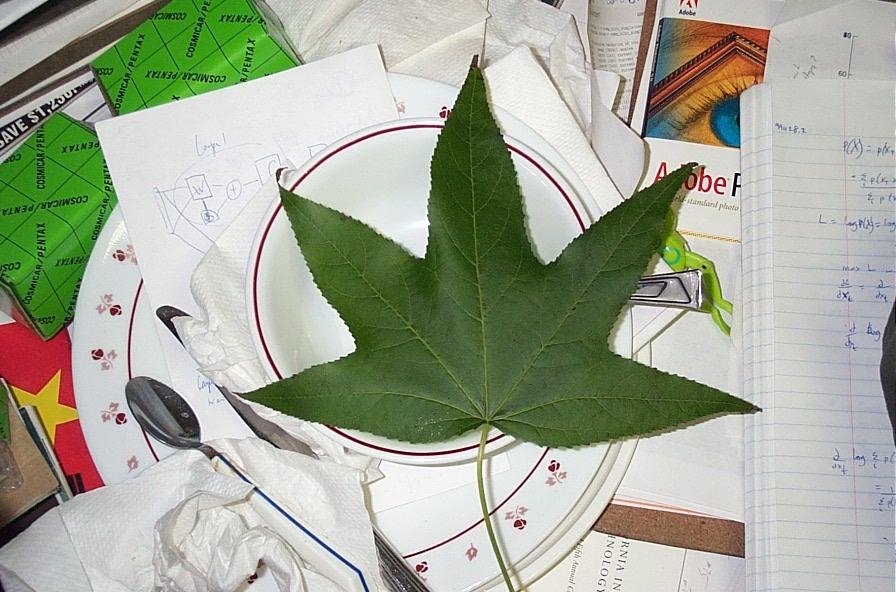} & \includegraphics[width=0.7in]{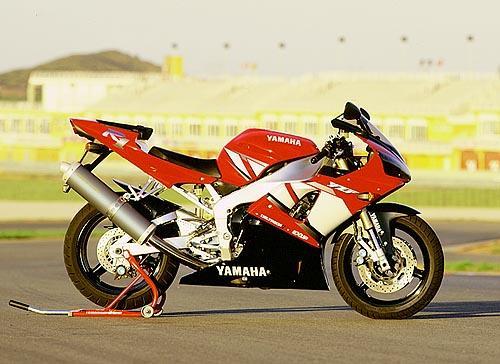} &
		\includegraphics[width=0.7in]{NearestImg/caltech5.jpg} & \includegraphics[width=0.7in]{NearestImg/caltech5.jpg} & \includegraphics[width=0.7in]{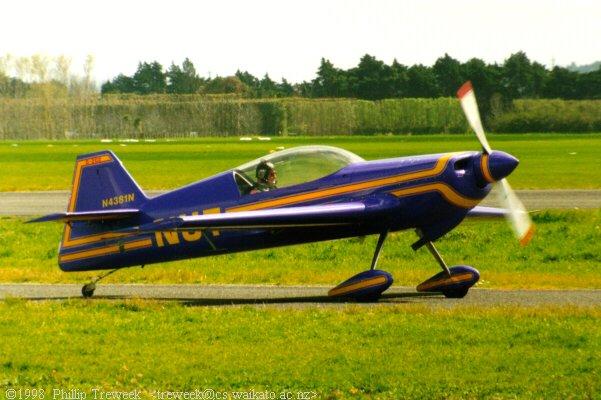}& \includegraphics[width=0.7in]{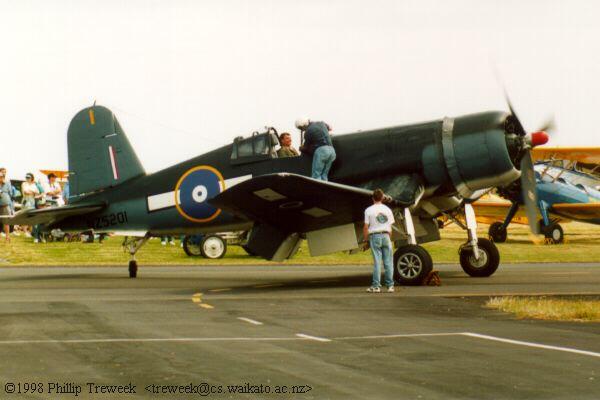}\\
		\includegraphics[width=0.7in]{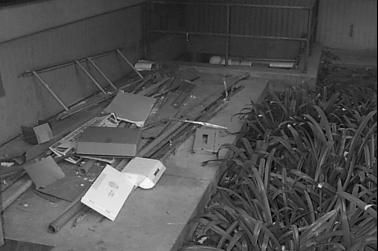} & \includegraphics[width=0.7in]{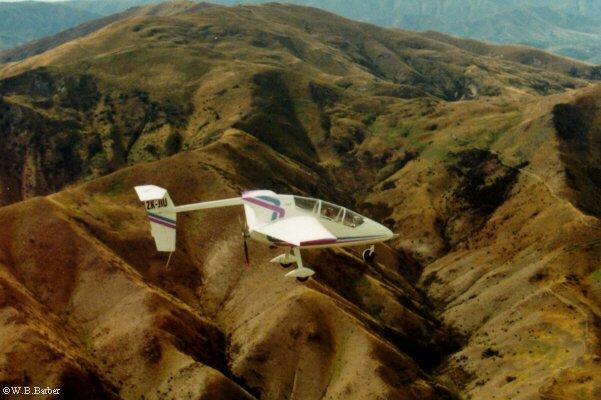} & \includegraphics[width=0.7in]{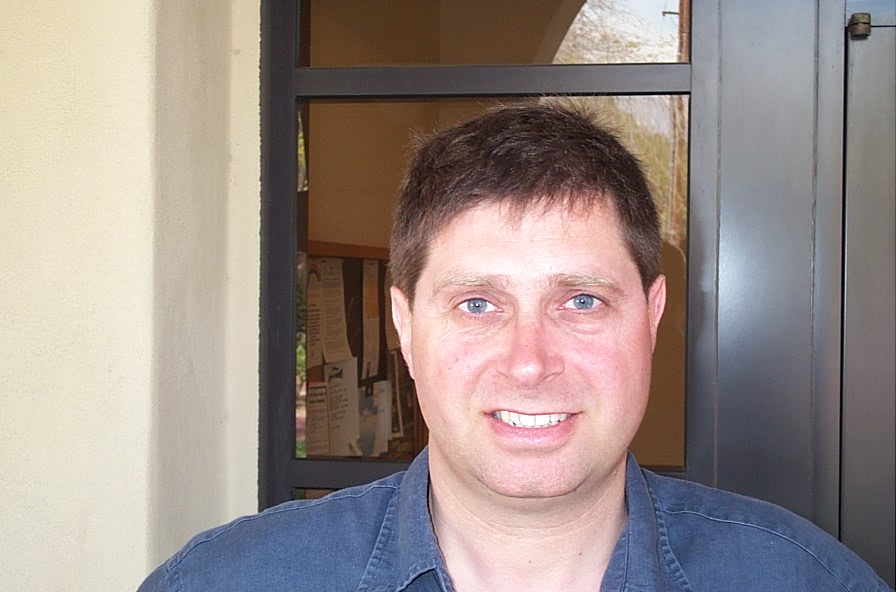} & \includegraphics[width=0.7in]{NearestImg/caltech213.jpg} &
		\includegraphics[width=0.7in]{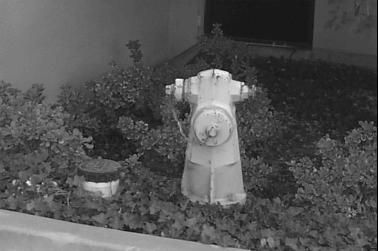} & \includegraphics[width=0.7in]{NearestImg/caltech21.jpg} & \includegraphics[width=0.7in]{NearestImg/caltech158.jpg}& \includegraphics[width=0.7in]{NearestImg/caltech21.jpg}\\
		\includegraphics[height=0.5in]{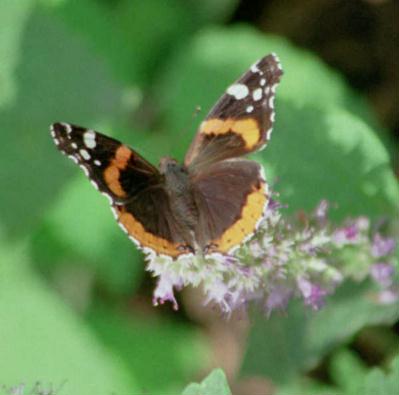} & \includegraphics[height=0.5in]{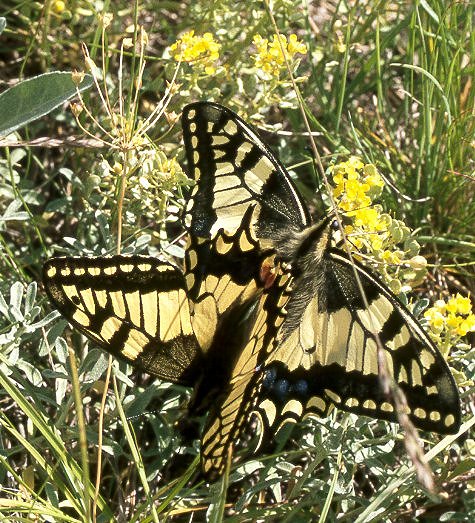} & \includegraphics[width=0.7in]{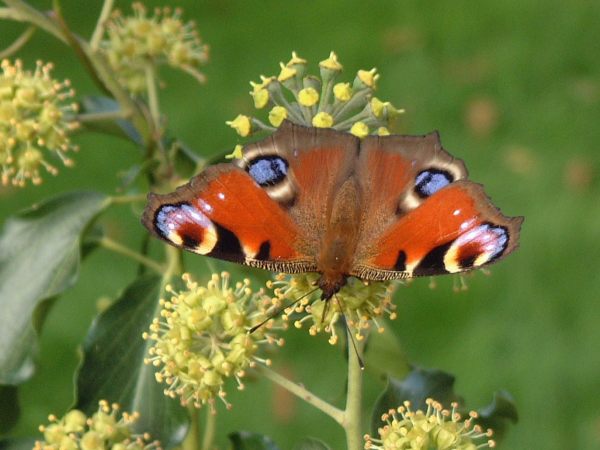} & \includegraphics[height=0.5in]{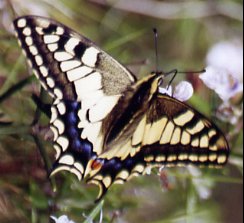} &
		\includegraphics[height=0.5in]{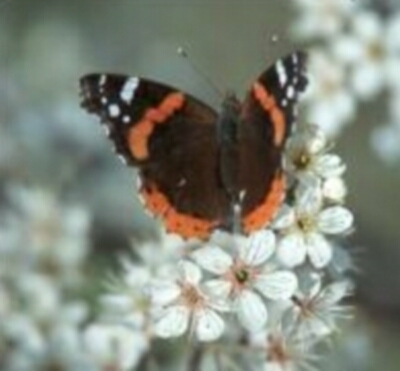} & \includegraphics[height=0.5in]{NearestImg/butterfly110.jpg} & \includegraphics[width=0.7in]{NearestImg/butterfly239.jpg}& \includegraphics[height=0.5in]{NearestImg/butterfly10.jpg}\\
		\includegraphics[height=0.5in]{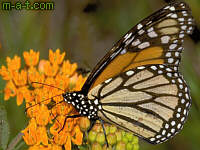} & \includegraphics[height=0.5in]{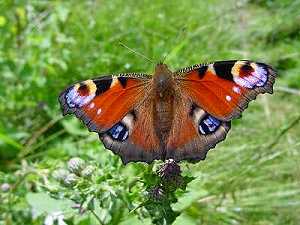} & \includegraphics[width=0.7in]{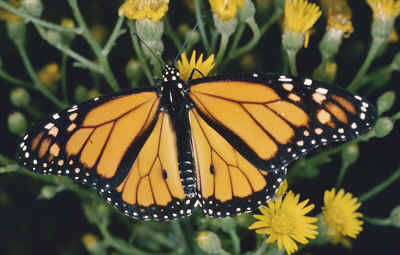} & \includegraphics[height=0.5in]{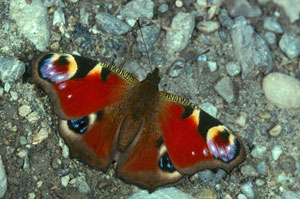} &
		\includegraphics[height=0.5in]{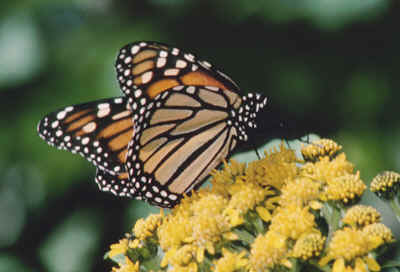} & \includegraphics[height=0.5in]{NearestImg/butterfly217.jpg} & \includegraphics[width=0.7in]{NearestImg/butterfly169.jpg}& \includegraphics[height=0.5in]{NearestImg/butterfly125.jpg}\\	
	\end{tabular}
\end{table*}

\subsubsection{Influence of parameters}
\begin{figure*}
	\centering
\subfigure[Corel]{
	\includegraphics[width=2.0in]{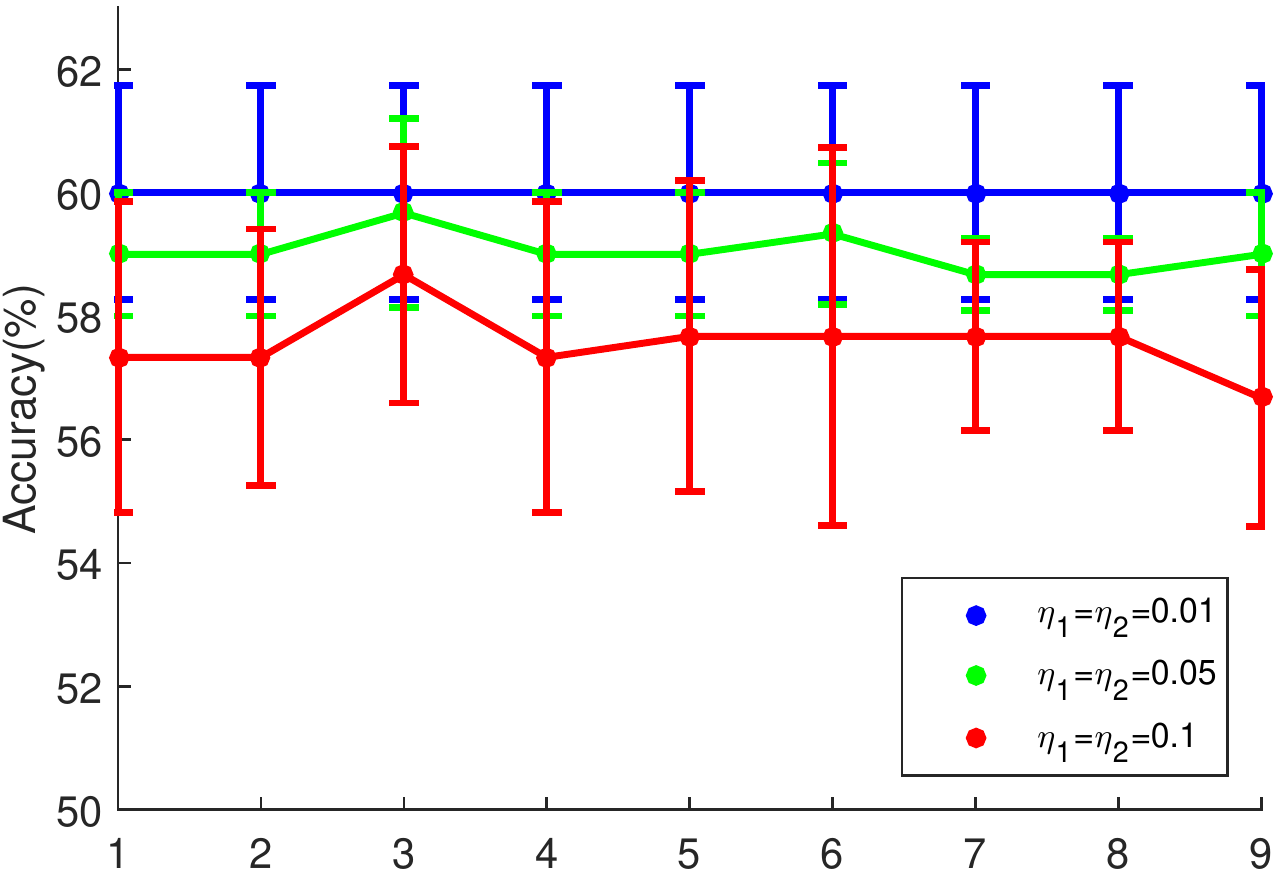}
}
\subfigure[Caltech]{
	\includegraphics[width=2.0in]{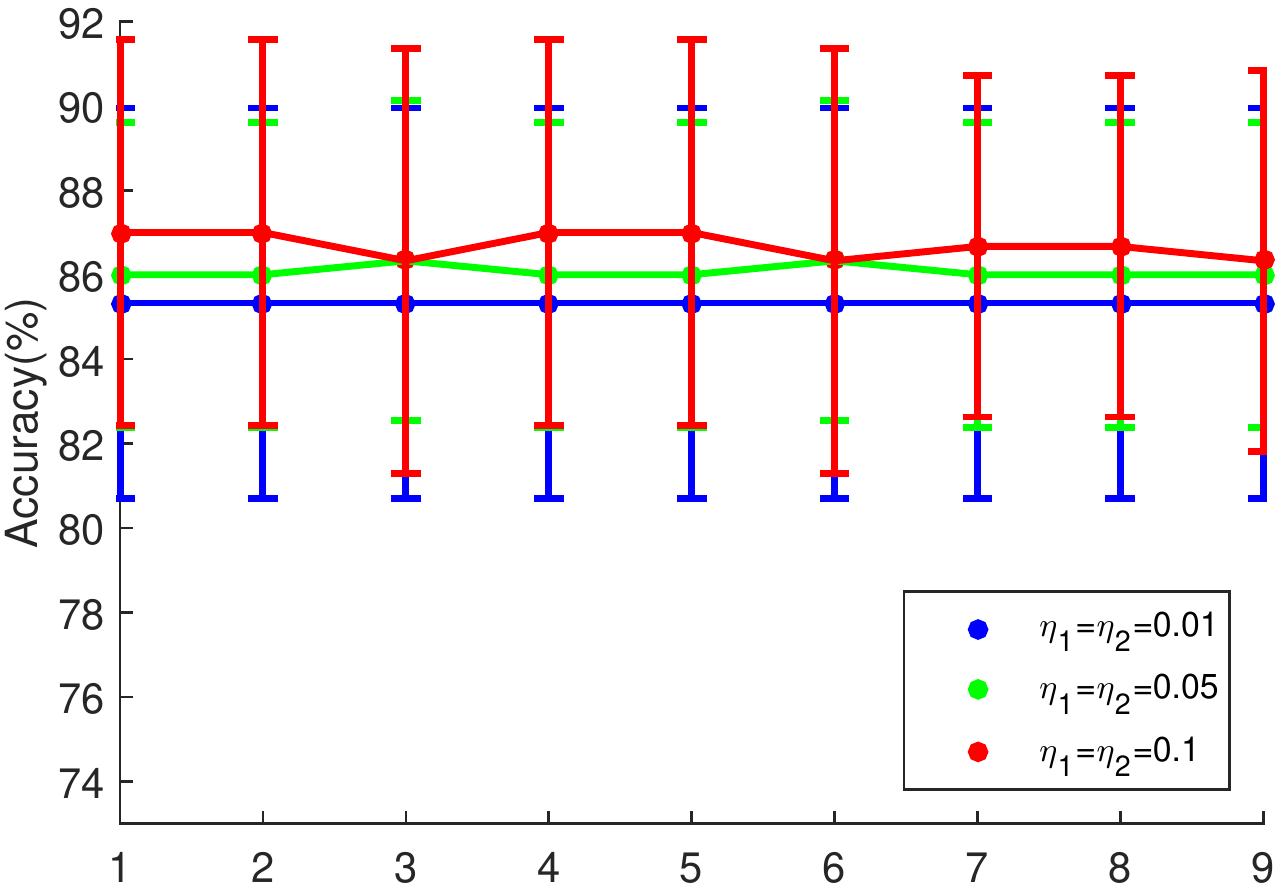}
}
\subfigure[Bike]{
	\includegraphics[width=2.0in]{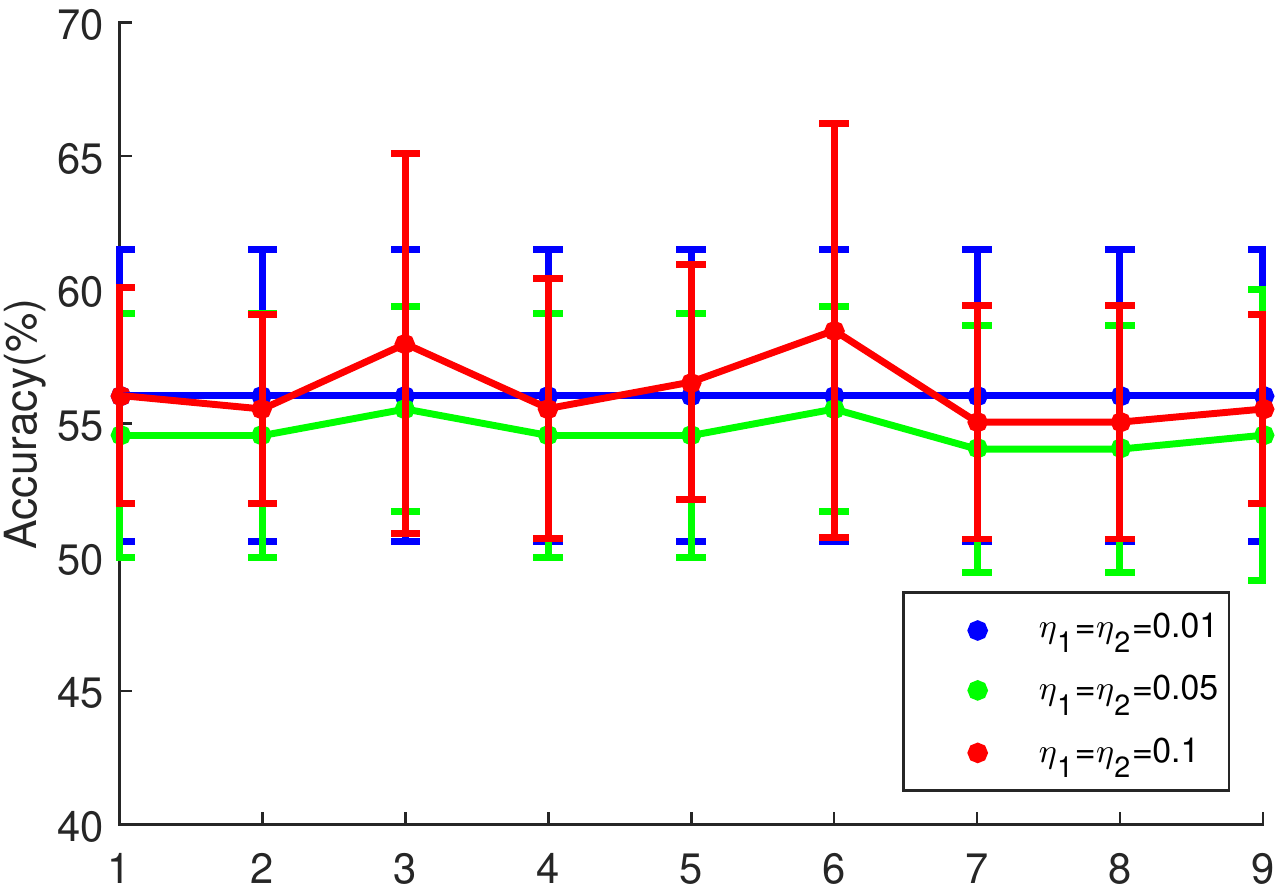}
}
\subfigure[Car]{
	\includegraphics[width=2.0in]{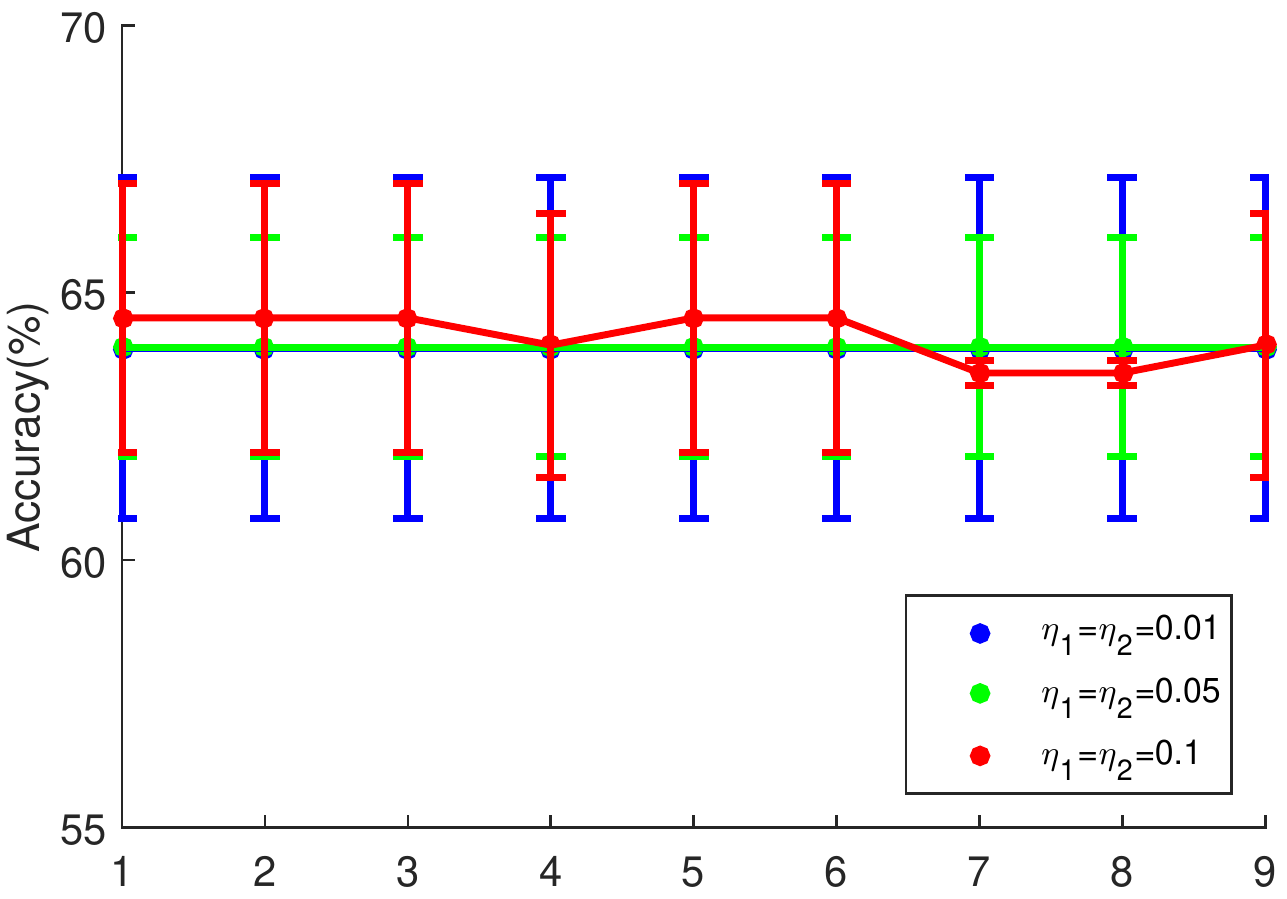}
}
\subfigure[Butterfly]{
	\includegraphics[width=2.0in]{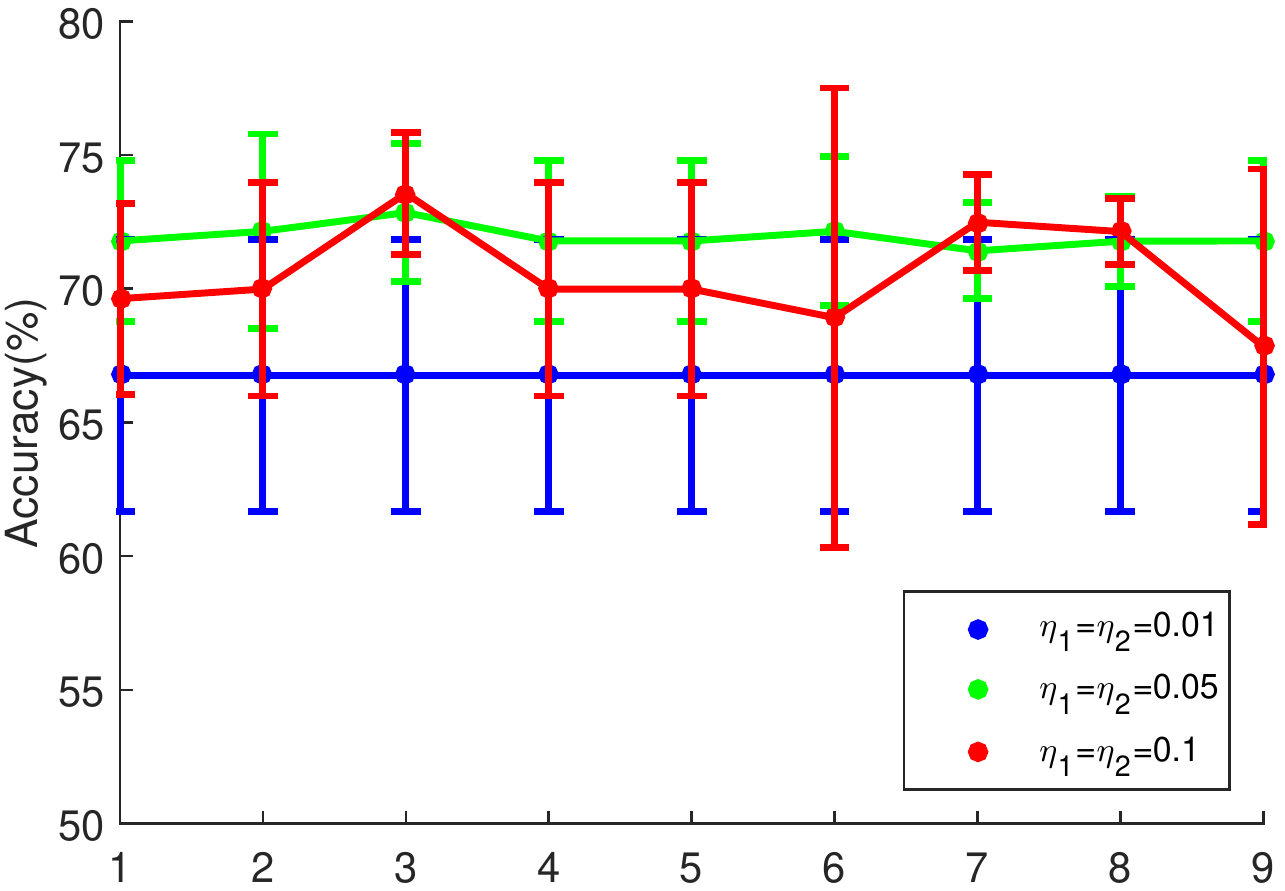}
}
\subfigure[Galaxy]{
	\includegraphics[width=2.0in]{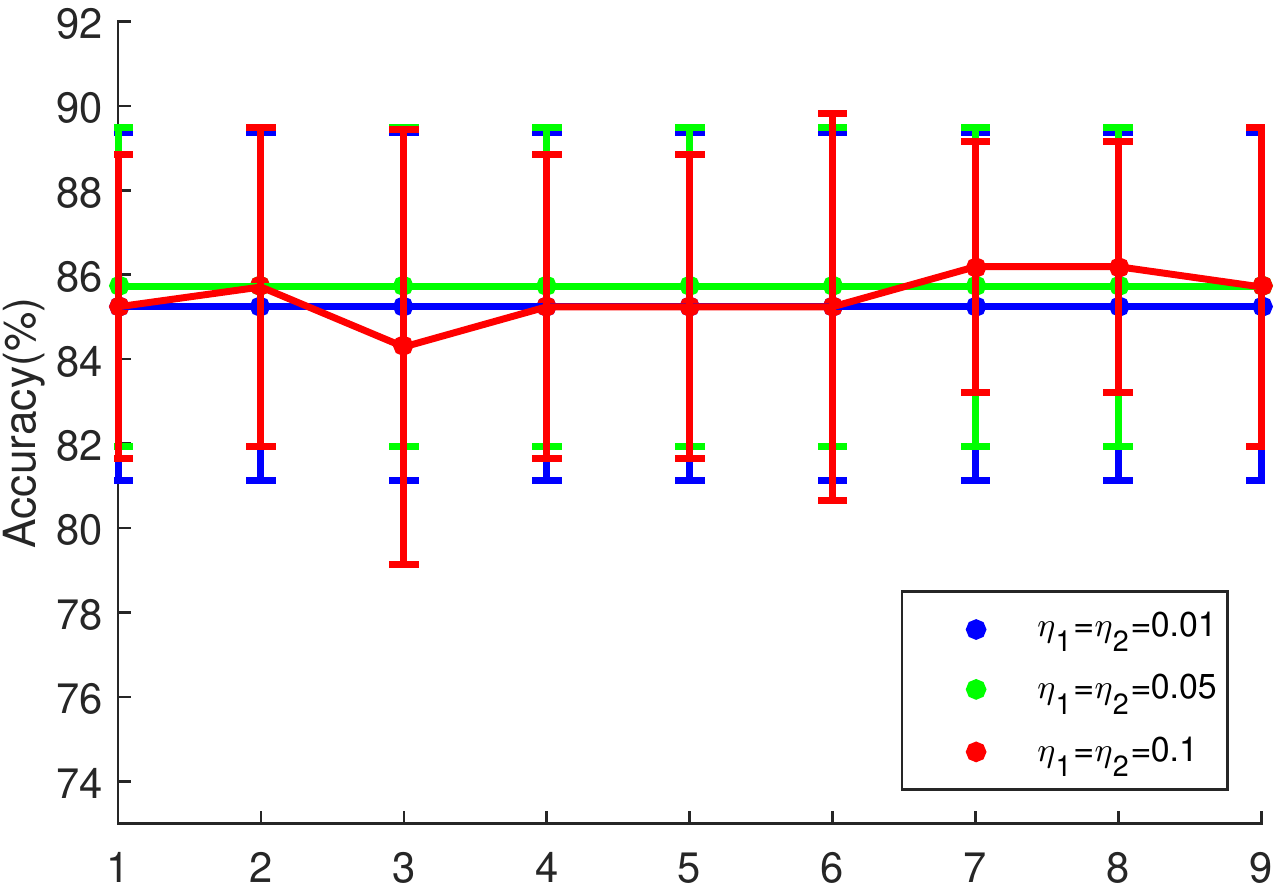}
}
\caption{Influence of the penalty parameters and learning rates.}
\label{fig:paras}
\end{figure*}

In this subsection, the influence of parameters $ \lambda, \mu $ and learning rates $ \eta_1, \eta_2 $ are explored.
The iteration numbers $ \tau, R $ are the same as the above experiments. As previously mentioned, $ \lambda $ and $ \mu $ are both selected from the set $ \{0.01, 0.1, 1\} $, so there are nine combinations for $ (\lambda, \mu) $, namely, $ \{(0.01,0.01),(0.01,0.1),(0.01,1),(0.1,0.01),(0.\\1,0.1),(0.1,1),(1,0.01),(1,0.1),(1,1)\} $. Fix the combination of $ \eta_1, \eta_2 $((0.01, 0.01),(0.05,0.05) or (0.1,0.1)), the accuracy curve and corresponding standard deviation with respect to the combination of $ \lambda, \mu $ can be got in Figure \ref{fig:paras}. Each subfigure corresponds to a selected dataset. The axis of $ x $ denotes nine combinations of $ \lambda, \mu $. For every combination of $ \eta_1, \eta_2 $, the accuracy fluctuates slightly, indicating that our model is robust to the penalty parameters when they are sufficiently small. For each combination of $ \lambda, \mu $, the accuracy does not change drastically with respect to $ \eta_1, \eta_2 $, which suggests the insensitivity of the method to the learning rates. The faint influence of parameters verifies the consistency of our new model.

\subsubsection{Influence of instance number}
\begin{figure*}
	\centering
	\subfigure[Corel]{
		\includegraphics[width=1.7in]{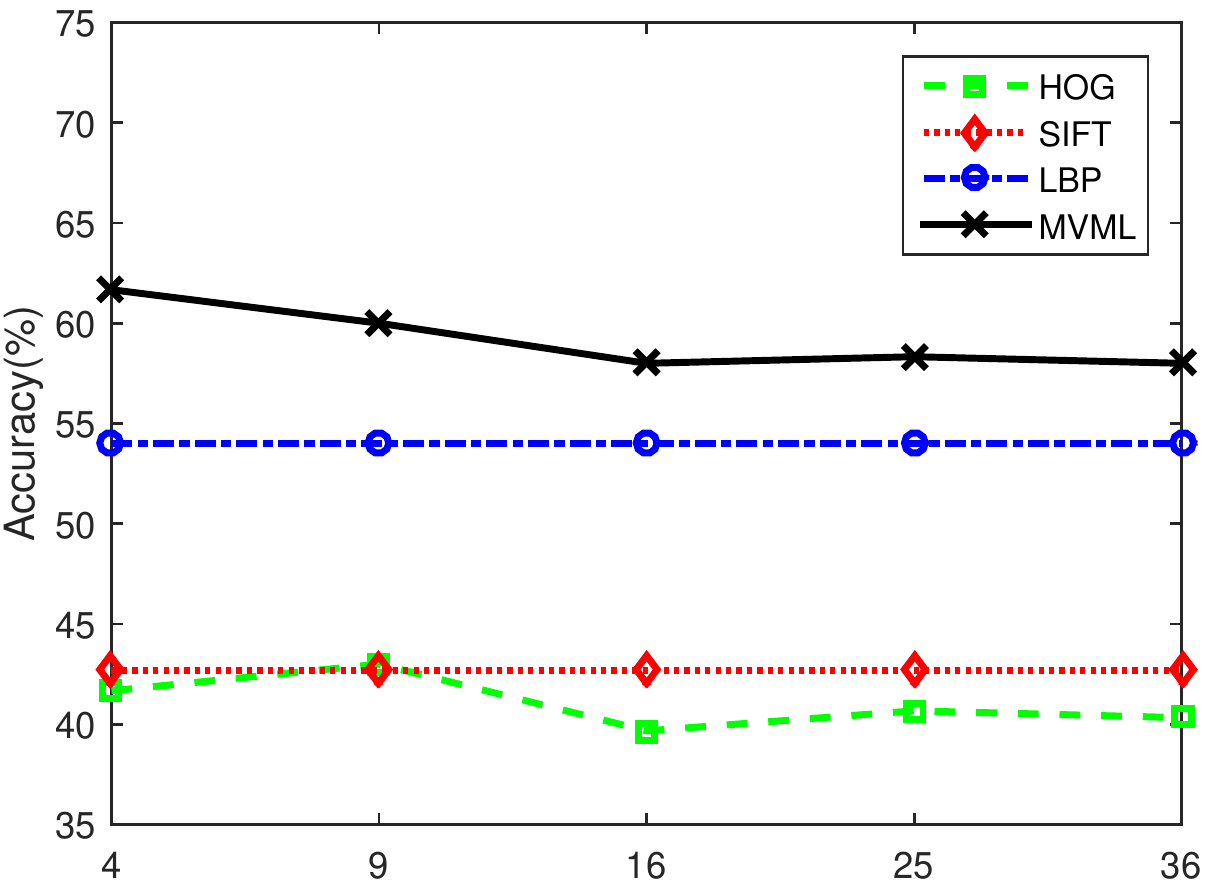}
	}
	\subfigure[Caltech]{
		\includegraphics[width=1.7in]{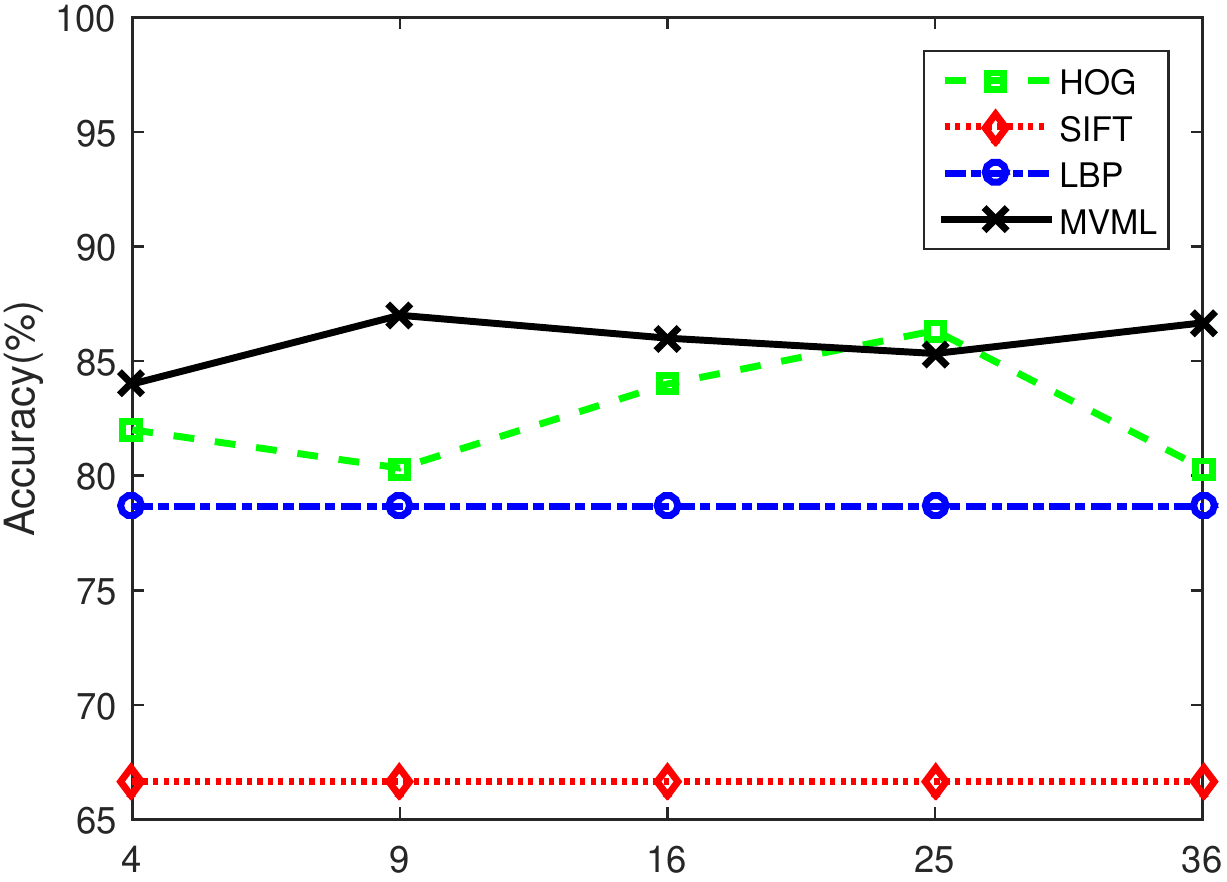}
	}
	\subfigure[Butterfly]{
		\includegraphics[width=1.7in]{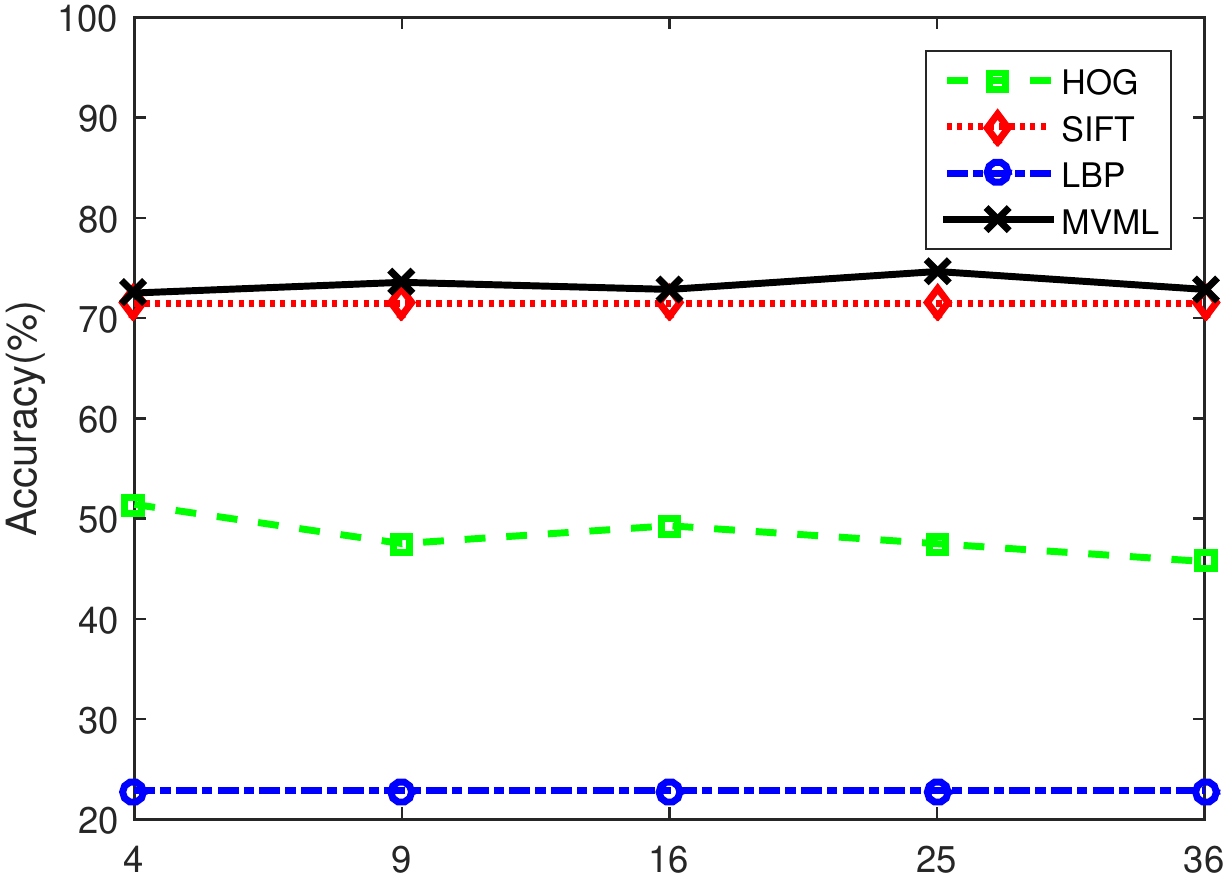}
	}
	\subfigure[Galaxy]{
		\includegraphics[width=1.7in]{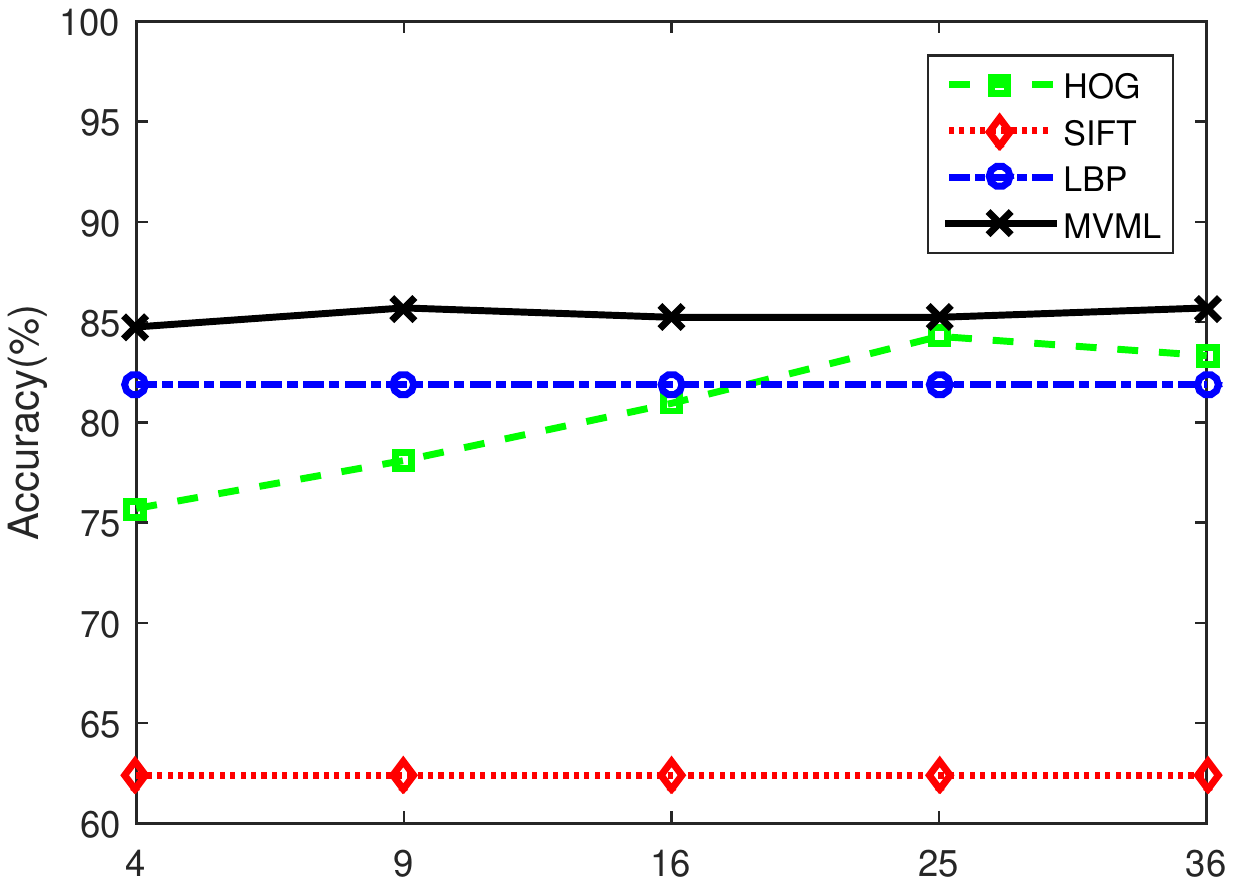}
	}
		\subfigure[Corel]{
			\includegraphics[width=1.7in]{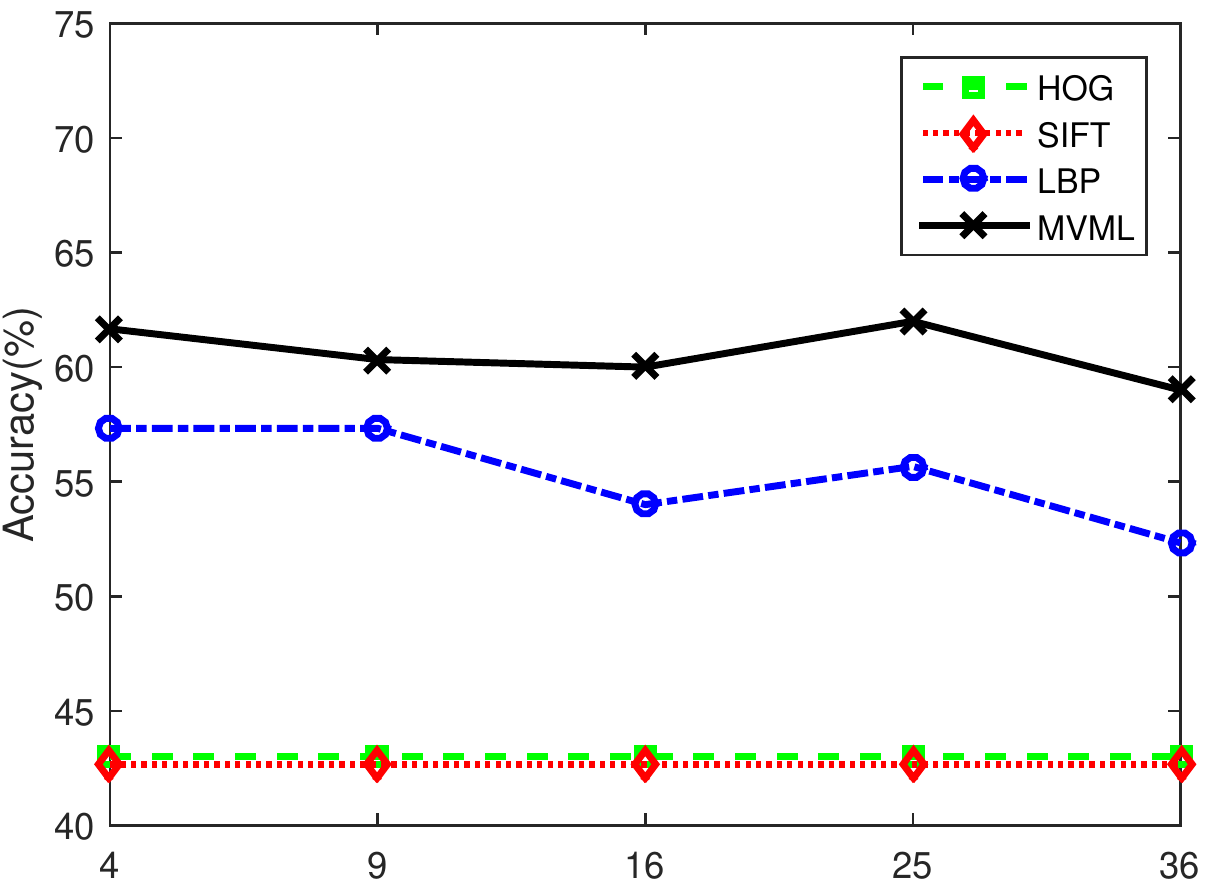}
		}
		\subfigure[Caltech]{
			\includegraphics[width=1.7in]{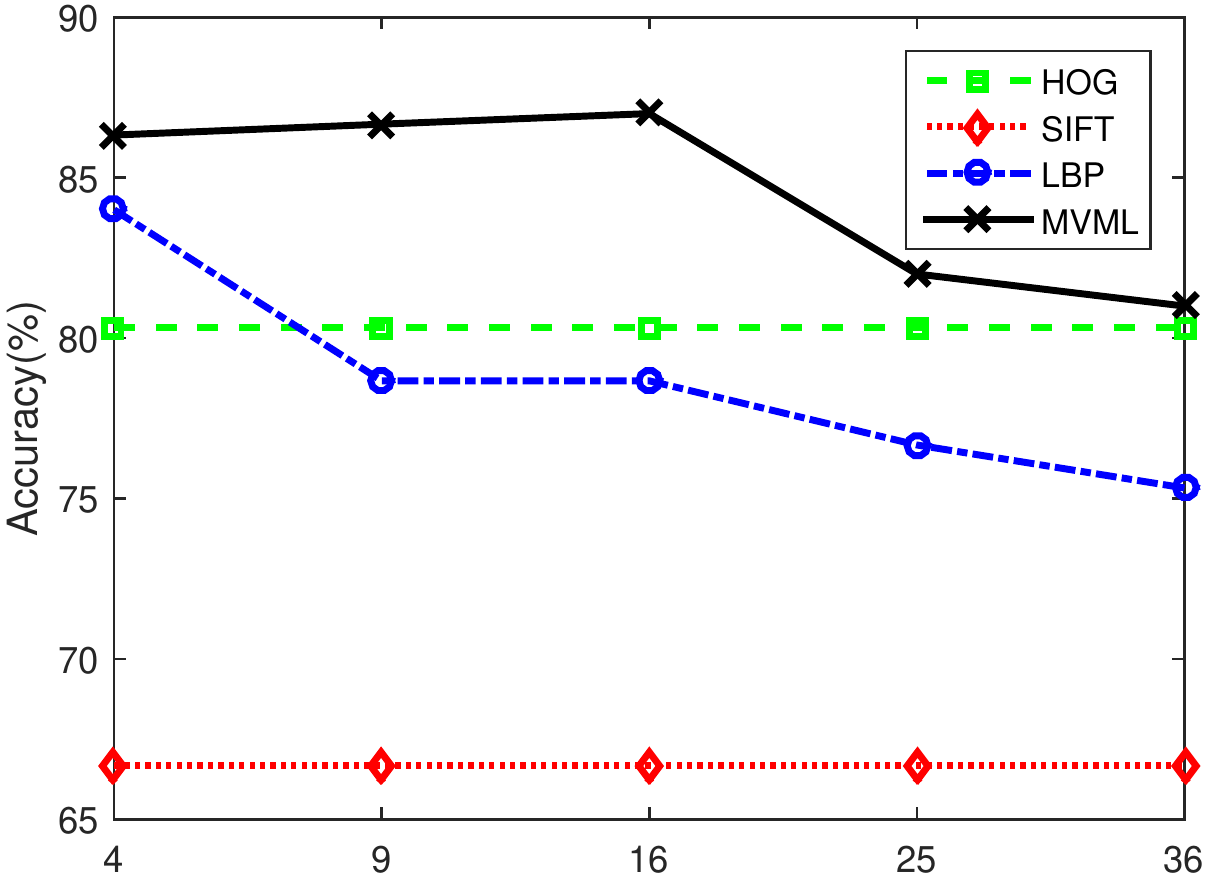}
		}
		\subfigure[Butterfly]{
			\includegraphics[width=1.7in]{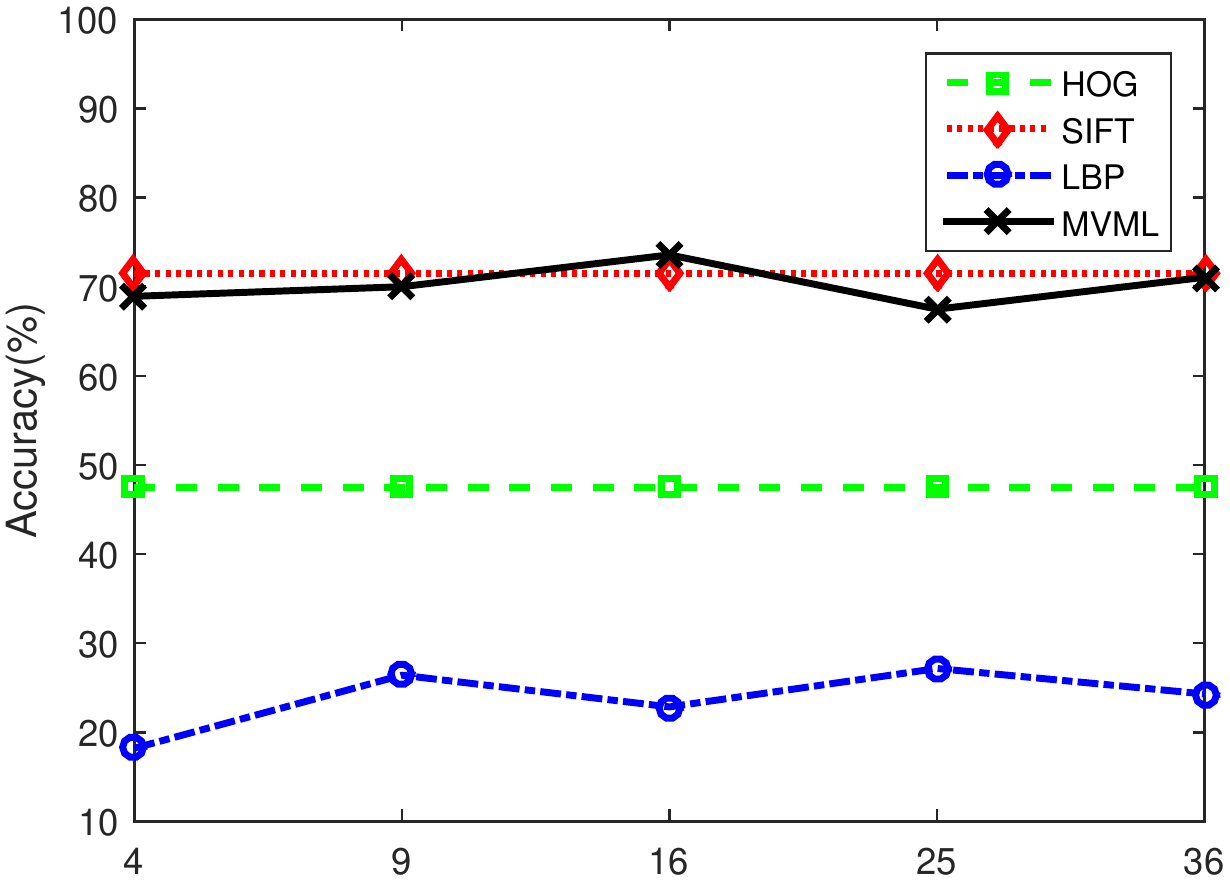}
		}
		\subfigure[Galaxy]{
			\includegraphics[width=1.7in]{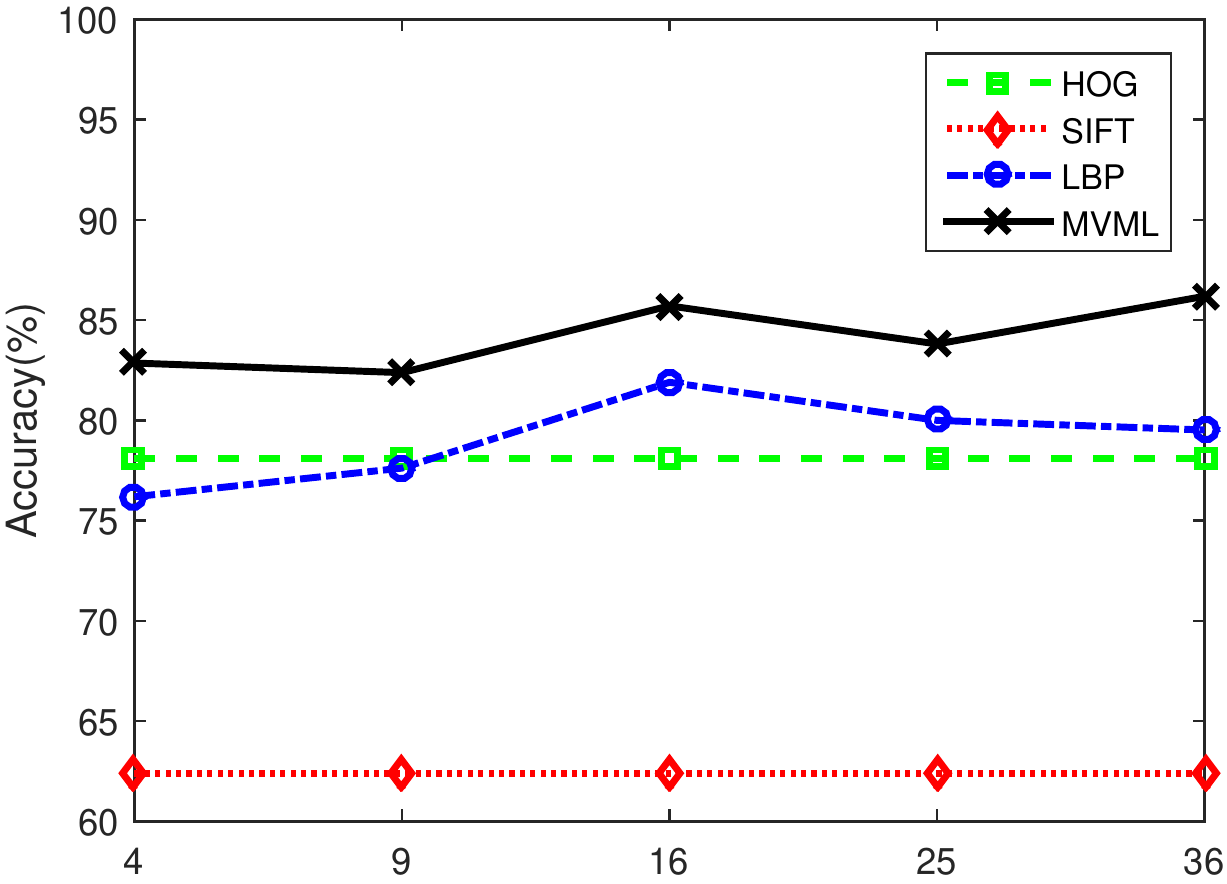}
		}
	\caption{Influence of instance number in HOG and LBP feature. In all the subfigures, the instance numbers of SIFT feature are the same as previous classification.In (a)(b)(c)(d), the instance numbers of LBP feature are all fixed as 16. In (e)(f)(g)(h), the instance numbers of HOG feature are all fixed as 9.}
	\label{fig:inst}
\end{figure*}

In the previous feature extraction, HOG and LBP are extracted as bag-of-words representation by dividing every image uniformly. In HOG and LBP feature, every image contains 9 and 16 instances respectively. Next we will investigate the influence of instance number in each bag. The penalty parameters $ \lambda, \mu $ are both set to be 0.01 and the learning rates $ \eta_1, \eta_2 $ are both set to be 0.1.
For HOG feature, each image is further divided into 4, 16, 25, 36 cells and the accuracy curve of single view and multi-view(H\&S\&L) with respect to instance number is depicted in Figure \ref{fig:inst}.(a)(b)(c)(d). It can be seen that the accuracy of single view changes slightly and the accuracy of multi-view is very stable, verifying the robustness of our model to instance number of HOG feature. The model can extract information consistently from HOG feature, immune to the instance number. For LBP feature, each image is further divided into 4, 9, 25, 36 patches and the accuracy curve of single view and multi-view(H\&S\&L) with respect to instance number is depicted in Figure \ref{fig:inst}.(e)(f)(g)(h). The accuracy curve of multi-view(H\&S\&L) has similar fluctuation trend with the accuracy curve of single view, but both curves have small amplitudes. The experiments demonstrate that our model is not sensitive to instance number, resulting from that a bag with different instance numbers can be actually seemed as feature extraction in different scale, which will not affect metric learning much.

\section{Conclusions}
\label{sec:conclu}	
In this paper, we propose a novel metric learning method for image classification, introducing metric learning technique into multi-view multi-instance task. For every image, feature extraction is conducted for three views, HOG, SIFT and LBP, to represent the image in complementary sides. For all the three visual features, every image  is presented by a bag, consisted of multiple instances. To unify the information of multiple views, the following efforts are made to make the classification performance as good as possible. First, a new distance function is designed for bags by calculating the weighted sum of the distances between instances. The distance between images with multiple views sums the weighted bag distance. 
Then we construct an optimization problem under probability framework to combine multi-view information, aiming to maximizing the joint probability that every image is similar with its nearest image.
The technique of metric learning is embedded to adjust the distance between instances from different bags.
The model can be iteratively solved by gradient descent and positive semi-definite projection alternately.
In future work, we will explore more kinds of features and design more efficient algorithm to solve our method due to its high computational cost.
\\

\textbf{Acknowledgement}\\
\indent This work has been partially supported by grants from National Natural Science Foundation of China (Nos .61472390, 11271361, 71331005, and 11226089), Major International (Regional) Joint Research Project (No. 71110107026) and the Beijing Natural Science Foundation (No.1162005).

\label{sect:bib}
\small
\bibliographystyle{plain}
\bibliography{MVMLbib}

\end{document}